\documentclass[runningheads]{llncs}
\pdfoutput=1
\usepackage{graphicx}
\usepackage{comment}
\usepackage{amsmath,amssymb} 
\usepackage{color}

\usepackage[mode=buildnew]{standalone}
\usepackage{array}
\usepackage{times}
\usepackage{epsfig}
\usepackage{soul}
\usepackage{amsbsy}
\usepackage{xcolor}
\usepackage{breqn}
\usepackage{graphicx}
\usepackage{array}
\newcolumntype{C}[1]{>{\centering\let\newline\\\arraybackslash\hspace{0pt}}m{#1}}
\begin{document}
\pagestyle{headings}
\mainmatter
\def\ECCVSubNumber{7301}  

\title{Giving Commands to a Self-driving Car: A Multimodal Reasoner for Visual Grounding} 


\titlerunning{Giving Commands to a Self-driving Car}
%
\author{Thierry Deruyttere \and
Guillem Collell \and
Marie-Francine Moens}
\authorrunning{T. Deruyttere et al.}
%
\institute{KU Leuven, Leuven, Belgium
\email{\{thierry.deruyttere,guillem.collelltalleda,sien.moens\}@kuleuven.be}}
\maketitle

\begin{abstract}
We propose a new spatial memory module and a spatial reasoner for the Visual Grounding (VG) task.
The goal of this task is to find a certain object in an image based on a given textual query.
Our work focuses on integrating the regions of a Region Proposal Network (RPN) into a new multi-step reasoning model which we have named a Multimodal Spatial Region Reasoner (\texttt{MSRR}).
The introduced model uses the object regions from an RPN as initialization of a 2D spatial memory and then implements a multi-step reasoning process scoring each region according to the query, hence why we call it a multimodal reasoner. 
We evaluate this new model on challenging datasets and our experiments show that our model that jointly reasons over the object regions of the image and words of the query largely improves accuracy compared to current state-of-the-art models.


\end{abstract}


\section{Introduction}
\begin{figure}[t]
  \centering
    \includegraphics[width=0.6\textwidth]{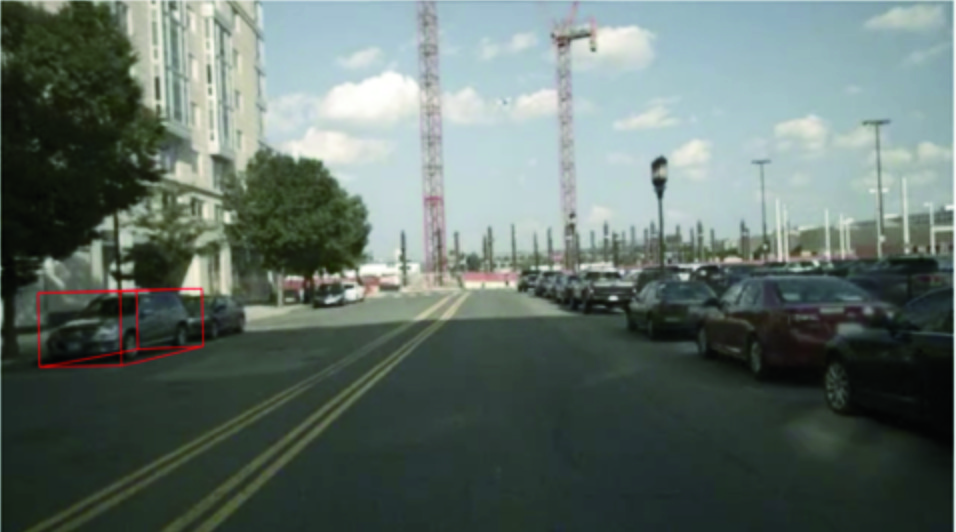}
\caption{Example command for a self-driving car. \textit{Command}: Turn around and park in front of \textbf{that vehicle in the shade}. The referred object is indicated with the red bounding box in the image and in bold font in the text.}
\label{fig:example-fig}
\end{figure}

Visual Grounding (VG) is a task relevant to many real-world scenarios and it is defined as follows:
Given a natural language expression, localize an image region based on this expression \cite{yu2018mattnet,mao2016generation}.
This task is useful for a variety of reasons. 
For instance, when taking a ride in a self-driving car, the passenger might want to instruct the car by saying, 
e.g.,``stop next to my friend with his red shirt next to the tree'' (Figure \ref{fig:example-fig}).
Another useful application of this task is Web search.
A user who is browsing a website that showcases different interiors of homes may issue the query: 
``I really like this blue lamp on the right side of the bed, can you check where it's from?''. 
The system can first locate the referred object in the image and then match the found object
with a database of retailers and similar objects for sale. 
When multiple instances of the object are present in the image, the language utterance may further constrain the search to the particular object described.
Correct resolution of the referred object demands joint reasoning over the query text, the image and its objects.

In this paper, we propose a novel method for the VG task that incorporates both the region ranking paradigm and the multi-step reasoning paradigm. 
To this end, we have created a new type of module, the \texttt{Spatial} module, that incorporates 2D\footnote{This can be potentially extended to 3D if suitable training data are available.} spatial information from extracted object regions in a spatial tensor which we call the \texttt{SpatialMap}. This module is integrated into a new multimodal model called \texttt{MSRR} which jointly reasons over the words of the query and object regions in the image. 
We evaluate this model on the Talk2Car dataset \cite{deruyttere2019talk2car} which is a referring expression dataset that contains referring commands given to self-driving cars. This dataset consists of multiple modalities (LIDAR, RADAR, Video, ...) but in this paper we only focus on the images and the referring expressions.
We furthermore validate our 
model on the well-known RefCOCO dataset \cite{yu2016modeling}. The main contributions of this paper are as follows:
\begin{enumerate}
 
    \item 
    We propose a novel integration method, \texttt{MSRR},  
    that decomposes a query in a multi-step reasoning process while continuously ranking 2D image regions during each step leading to low-scoring regions to be ignored during the reasoning process.
    This process leads to a better and transparent coupling between region proposals and the decomposed queries.

    
    \item We introduce a new \texttt{Spatial} module (used by \texttt{MSRR}), 
    which stores 2D spatial information in a tensor called a \texttt{SpatialMap}. 
    
    \item We evaluate our model on the real-world and non-curated Talk2car \cite{deruyttere2019talk2car} dataset and show that our results improve the best state-of-the-art model by almost 10\% in terms of IoU of the found referred object.
    
    \item We also show that using spatial information in the \texttt{Spatial} module vastly improves performance over an ablated model which does not use this information.

\end{enumerate}

\section{Related Work}
\begin{figure*}[t!]
  \centering
    \includegraphics[width=1\textwidth]{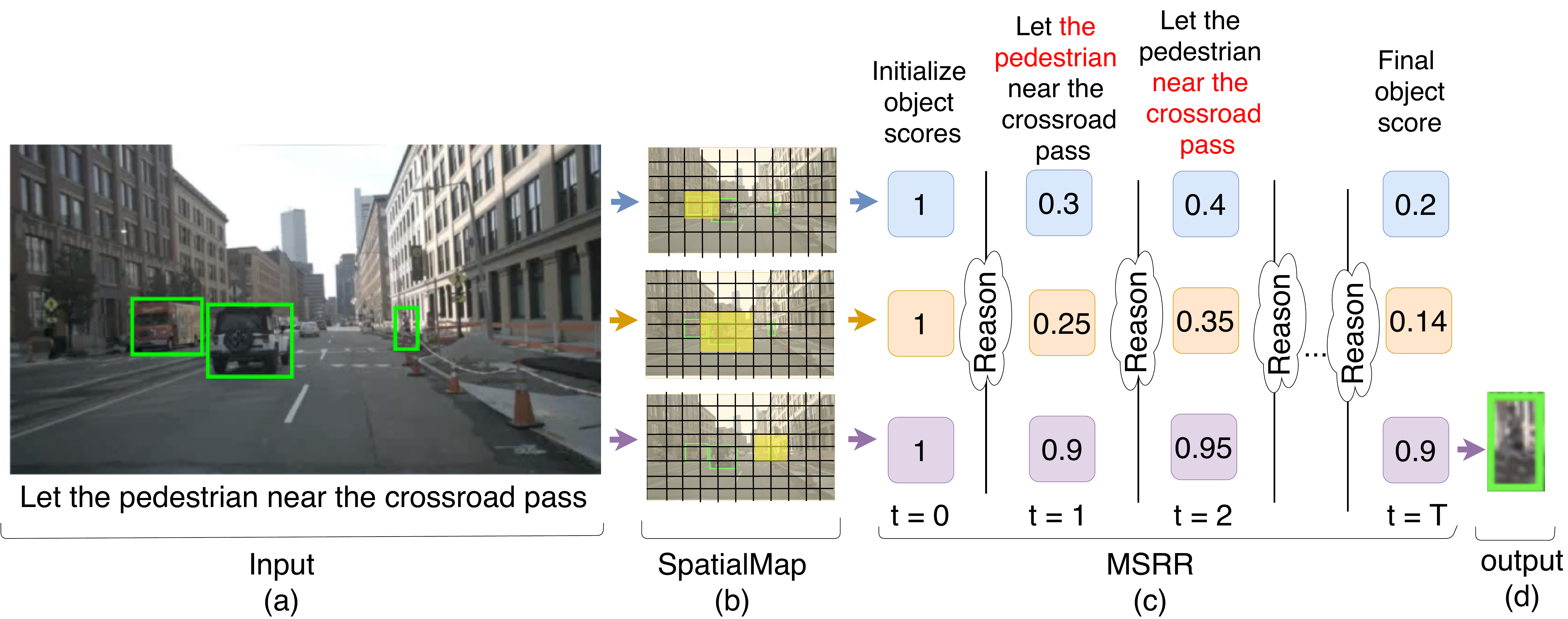}
\caption{
Sketch of \texttt{MSRR}. In step (a) an image is given together with a command and $O$ (in this case $O=3$) object regions. In step (b), each of these regions receives an entry in the 
tensor called the \texttt{SpatialMap}. Step (c) represents the reasoning process of our model. At the start ($t=0$), each region has a score of 1.
In subsequent reasoning steps, regions receive new scores based how well they align with the words that are focused in that reasoning step. These focused words are indicated in red in each step. At the end of the reasoning ($t=T$), the region with the highest score is returned as the answer of the model (step (d)).
}
\label{fig:gps-idea}
\end{figure*}

\paragraph{Visual Grounding}
The Visual Grounding (VG) task is 
defined as follows: given a natural language expression (query), localize the image region relevant to this description \cite{yu2018mattnet,mao2016generation}.
\cite{karpathy2014deep} were one of the first to 
score the regions extracted from a given image based on the inner product between the representations of said regions and the text expression.
\cite{mao2016generation,yu2016modeling} see VG as an image captioning task. They create captions for parts of the image to describe what is visible in a selected region. The regions are retrieved with a Region Proposal Network (RPN) and a caption is generated with an RNN. Finally, the generated captions are ranked according to their similarity with the query. The highest scoring caption is then selected as answer.
Attention mechanisms, which have 
been proven successful in other tasks such as VQA (see below), have also been investigated. \cite{yu2018mattnet,liu2019improving} 
leverage attention to decompose the referring textual expression into three modular components to aid detection of subject, location, and relationship to other objects in the image.
\cite{deng2018visual} create three different modules to focus on the query, image and objects. 
These modules share guidance vectors between each other in a circular manner to guide the reasoning process in multiple steps. This final model from \cite{deng2018visual} is included as a state-of-the-art baseline in our paper. 

\paragraph{Visual Question Answering}
Visual Question Answering (VQA) -- commonly seen as a visual Turing test -- is a related task that
requires answering both general and specific questions about a given image. There are two main approaches to the task.
 The first one consists in fusing the extracted features of the image together with the encoded 
 textual question \cite{FukuiPYRDR16,kafle2016answer,ZhouTSSF15}. 
 This fusion can be done in many different ways, including multiplication, addition and bilinear pooling.
 A second approach which has gained popularity over the last few years regards attention-based mechanisms steered by the analysis of the natural language question.  
For instance, \cite{Johnson2017,Suarez2018} have achieved state-of-the-art results on the CLEVR VQA dataset by using modular neural networks with attention.
In their approach, a LSTM network yields a syntactic parse of the question and outputs the layout of the modular network. \cite{hu2018explainable} avoid a syntactic parse by using a soft attention over the modules of the network based on a multi-step reasoning process over the question.
\cite{Hudson2018} have generalized the idea of multi-step reasoning and created a new type of memory module called Memory, Attention and Composition (\texttt{MAC}).
The \texttt{MAC} modules are chained together to form a multi-step reasoning process that has yielded promising results for VQA.
In contrast with ours, none of the above approaches \textit{combines} a multi-step reasoning process -- jointly over the natural language expression and the image -- with a spatial 2D map for the recognized objects. 

\paragraph{Spatial Memory Modules}
The textual queries in VG or VQA tasks often contain relationships between objects that constrain the referred object. Their processing often requires spatial reasoning with the information in the image as well as a memory to store the spatial layout of the image patterns in a 2D map. We also witness interest in using 2D maps in planning and navigation \cite{DBLP:conf/cvpr/GuptaDLSM17,DBLP:conf/iclr/ParisottoS18}. In \cite{DBLP:conf/iccv/ChenG17} the authors 
propose a type of spatial memory based on CNNs that encode instance-level spatial knowledge. In \cite{Henriques_2018_CVPR} visual observations obtained from a moving camera are stored with the help of an allocentric spatial memory module. The latter allows the network to understand the captured visual world independently of the observation point. The spatial map is a 
representation of an environment storing information that a deep neural network module learns to distill from RGB-depth input. However, none of the above works implement a multi-step joint reasoning over a language expression and spatial memory map, as is done in this paper. 

\section{MSRR Model}
\begin{figure*}
  \centering
    \includegraphics[width=1\textwidth]{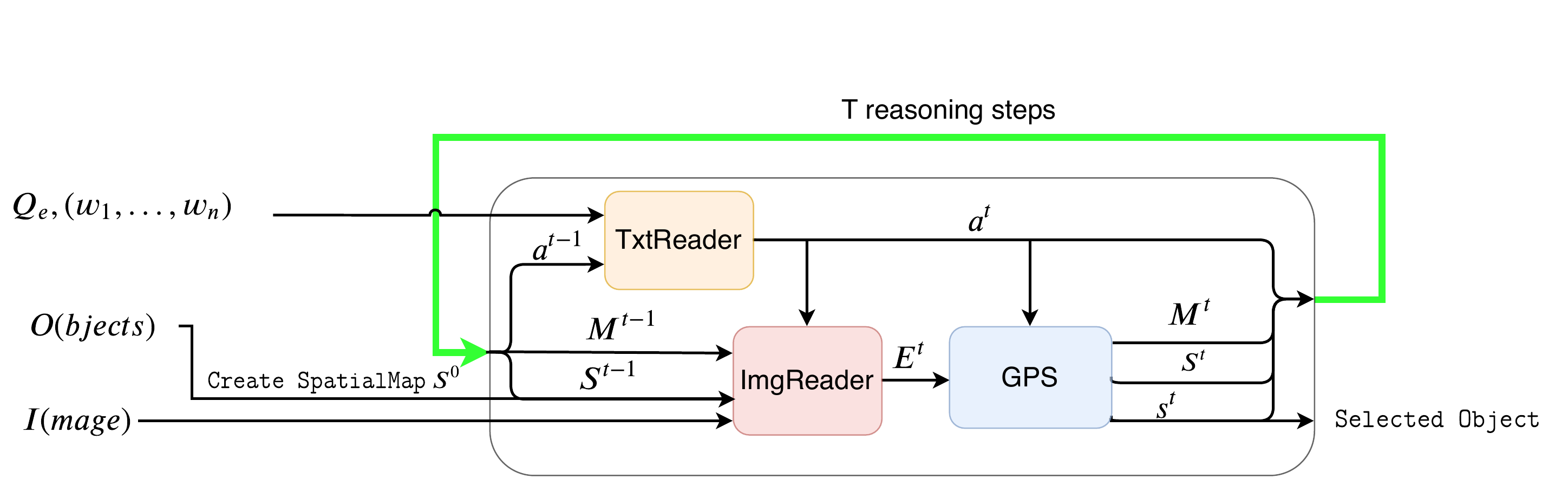}
\caption{The MSRR.}
\label{fig:gps-reasoner}
\end{figure*}

Before explaining the details of \texttt{MSRR}, we first describe the intuition behind  
the architecture.
The goal of the VG task is to find a region or object in an image that has the best alignment with the given query.
As the query can contain both \textit{absolute} spatial information (i.e., ``The man walking on the left'') or \textit{relative} information (i.e., ``The man next to the car'') the model should be able to cope with this. 
To solve this task, we 
first extract objects of interest and indicate their individual positions with a \texttt{SpatialMap}. 
We also assign a starting score of $1$ to each of these extracted objects.
The goal is to implement a model that reasons over the image, the regions and the query altogether such that it lowers the score of objects that do not seem important while on the other hand keeping the score of relevant objects high. Finally, the highest-scoring object is selected.

The above steps are executed in the following reasoning process (Figure \ref{fig:gps-idea}).
First, the model attends to certain words of the query that appear to be important to the current reasoning step.
Second, 
the model finds whether any regions 
correspond with these attended words or not. This is done by first imposing the spatial map of every object individually over the image, effectively telling the model where a certain object of interest lies. Then, with soft-attention, the model extracts data from the image that it deems relevant to the attended words and to the location of the object.
By using soft-attention we allow the model to look around 
the image, focusing on 
certain regions relative to the marked object.
This data is then encoded in an object-specific vector to represent all the extracted data for this object.
This vector is then used to calculate a score based on its alignment with the query.
This process is executed $T$ times, and the highest-scoring object is finally selected as answer.

This process is implemented through three different \textbf{modules} that interact with each other: (i) a \texttt{TxtReader} (section \ref{sect:txtreader}), based on \cite{deng2018visual,Hudson2018,yu2018mattnet,hu2018explainable}, that controls the decomposition of the query text and thus dictates how the reasoning process will unfold, (ii) a novel \texttt{Spatial} module (section \ref{sect:gps}) that functions as the 2D spatial memory of the model by using a \texttt{SpatialMap} (section \ref{sect:gps-map-closer}), (iii) an \texttt{ImgReader} (section \ref{sect:imgreader}), 
weakly inspired by \cite{deng2018visual,Hudson2018}, that extracts information from a given image based on the directions of (i) and (ii).
Our model with the interactions between 
modules is illustrated 
in Figure \ref{fig:gps-reasoner}.
By looping \texttt{TxtReader}, \texttt{Spatial} and \texttt{ImgReader} $T$ times
\texttt{MSRR} can perform a sequential reasoning process by using the output of the modules at step $t-1$ as input to the same modules at step $t$. Each iteration $t$ corresponds to one \textbf{reasoning step}.


Our model emphasizes the joint reasoning with both modalities and the transparency of the proposed novel alignment method between the query and objects in the image at each reasoning step. Furthermore, our model allows scoring the objects in parallel, which can substantially boost speed in scenarios where this is needed (e.g., a self-driving car interpreting commands).


As illustrated in Figure \ref{fig:gps-reasoner}, the input to 
\texttt{MSRR} are: (a) the extracted features of an image, 
$I \in \mathbb{R}^{H_f \times W_f \times d}$ with $H_f$ and $W_f$ 
the height and width of the feature map, respectively;
(b) the query encoded by a BiLSTM, $Q_e \in \mathbb{R}^d$, and the word embedding vectors, $(w_1,...w_n) \in \mathbb{R}^{n \times d}$, extracted from the same BiLSTM model; and (c) a \texttt{SpatialMap} $\in \mathbb{R}^{O \times H_f \times W_f}$ tensor (section \ref{sect:gps-map})
that has been initialized with $O$ object proposals previously extracted by a region proposal algorithm (e.g., an RPN).
The output of the model 
are the 2D image coordinates of an object region 
$[x, y, w, h]$ with $(x,y)$ denoting the top left corner and $(w,h)$ the width ($w$) and height ($h$) of this object.
\subsection{TxtReader}
\label{sect:txtreader}
The \texttt{TxtReader} module 
guides the reasoning process by selecting 
which words of the referring expression are the most important to focus on in each step $t$.
For instance, for the query ``\textit{man next to tree}'', the system might focus on ``\textit{man}'' in the first step, on ``\textit{next to}'' in the second step and on ``\textit{tree}'' in the final reasoning step.
Before describing this module and the subsequent ones we shall introduce \textbf{notation} first.
We use a superscript $t$ (e.g.,: $x^t$) to indicate that a 
variable belongs to 
reasoning step $t$. 
The inputs to this module are: (i) the query embedding $Q_e \in \mathbb{R}^d$, (ii) the embeddings of the $n$ query words $(w_1, ..., w_n) \in \mathbb{R}^{n \times d}$, and (iii) the embedding of the \textit{focused words} from step $t-1$ stored in the vector $a^{t-1} \in \mathbb{R}^d$.
Based on these inputs, the module calculates the new $a^{t} \in \mathbb{R}^d$ vector.
As a first step, \texttt{TxtReader} projects the query embedding $Q_e$ with a linear layer\footnote{As convention we describe the format of weight matrices as $\mathbb{R}^{output\_dim \times input\_dim}$. We also add superscripts $t$ to weight matrices to indicate 
if they are step-dependent or not. 
If not, 
they are reused across
steps. Integer subscripts after a comma simply index each different set of weights within a given module (e.g., ``txt'' $\sim$ \texttt{TxtReader}).}:
\begin{equation}
    q^t = W^t_{txt,1} Q_e+b,
\end{equation}
where $b$ is the bias and $W^t_{txt,1}$ $\in \mathbb{R}^{d \times d}$.
Then, the model combines $q^t$ with 
$a^{t-1} \in \mathbb{R}^d$ which  
weights the words from the previous step according to their importance:

\begin{equation}
    c^t = W_{txt,2}[q^t; a^{t-1}]+b,
    \label{eq:ctrl1}
\end{equation}
with [;] the concatenation operation and $W_{txt,2}$ $\in \mathbb{R}^{d \times 2d}$.
Subsequently, the model scores the $n$ word embeddings against $c^t$ and creates an attention vector $m^t \in \mathbb{R}^n$ based on this score:
\begin{equation}
    m^t = \texttt{softmax}(W_{txt,3}((w_1,...,w_n) \bullet c^t)+b),
    \label{eq:ctrl2}
\end{equation}
where $W_{txt,3} \in \mathbb{R}^{1 \times d}$ and $\bullet$ indicates that we broadcast the second term to the shape of the first one and then apply element-wise multiplication.
Finally, to get the embedding of the \textit{focused words} at step $t$, a weighted sum is applied over the 
word embeddings $(w_1, ..., w_n) \in \mathbb{R}^{n \times d}$ to get the
vector $a^t \in \mathbb{R}^{d}$. 
This is also the output of this module.
\begin{equation}
    a^t = \sum_{i=1}^n m^t_i w_i
    \label{eq:ctrl3}
\end{equation}

\subsection{ImgReader}
\label{sect:imgreader}
The \texttt{ImgReader} module is responsible for extracting data from the image based on both the focused words from the \texttt{TxtReader} and 
the object-specific data from the \texttt{Spatial} module (section \ref{sect:gps}).
The inputs to this module are: (a) the embedding of the \textit{focused words} $a^t \in \mathbb{R}^d$,
(b) the image features $I \in \mathbb{R}^{H_f \times W_f \times d}$, (c) a matrix $M^{t-1} \in \mathbb{R}^{O \times d}$ which stores the 
previous memory for the $O$ objects (section \ref{sect:update-spatial-mem}),
and (d) the \texttt{SpatialMap} tensor $S^{t-1} \in \mathbb{R}^{O \times H_f \times W_f}$ 
of all objects (section \ref{sect:gps-map}). 
For clarity, the functionality of this module is split into two. 
The first function of the module is to indicate the spatial location and the importance of an object (section \ref{sect:spatial-loc}).
Its second function is to extract information based on this first step (section \ref{sect:extract-info}). 

\subsubsection{Spatial Location and Importance of an Object}
\label{sect:spatial-loc}
In order to indicate both the location and the importance of an object $o$, the image features $I \in \mathbb{R}^{H_f \times W_f \times d}$ are multiplied element-wise with the \texttt{SpatialMap} tensor $S^{t-1}_o \in \mathbb{R}^{H_f \times W_f}$ (section \ref{sect:gps-map}) for this object $o$.
This returns a spatially-weighted image tensor, $I_{o}^t \in \mathbb{R}^{H_f \times W_f \times d}$, where the features that belong to grid cells that fall within the object region $o$ are more visible than the rest.
\begin{equation}
\label{eq:mult-spatial}
    I_{o}^t = I \bullet S^{t-1}_{o} 
\end{equation}

\subsubsection{Extracting Information}
\label{sect:extract-info}
Now that the model knows where an object is located 
and how important said object is, 
it can extract information from the image. 
To do this, it 
uses the 
image tensor $I_{o}^t$ (eq. \ref{eq:mult-spatial}) in addition to the previously extracted image data for the current object, $M_o^{t-1} \in \mathbb{R}^{d}$  (section \ref{sect:gps}), and the words that are important in this reasoning step, $a^t$, in the following manner.
First, in order to find out which information in the image may be important according to the 
memory data of this object, $M_o^{t-1}$, 
we integrate $M_o^{t-1}$ with $I_{o}^t$ in $Z_{o}^t \in \mathbb{R}^{H_f \times W_f \times d}$:
\begin{equation}
	Z_{o}^t = (W_{ir,1} I_{o}^t +b) \bullet (W_{ir,2} M_{o}^{t-1} +b),
\end{equation}
with $W_{ir,1} \in \mathbb{R}^{d\times d}$ and $W_{ir,2} \in \mathbb{R}^{d\times d}$.
As the model now knows what should be important in light of the previously extracted data, the only thing that remains 
is to score each feature cell in $Z_{o}^t$ according to the focused words $a^t \in \mathbb{R}^d$ of this reasoning step in order
to calculate a softmax distribution over the 2D grid (of image features):

\begin{equation}
\label{eq:totalImp-score}
    D_{o}^t = \texttt{softmax}(W_{ir,3} (Z_{o}^t \bullet a^t) + b),
\end{equation}
where $W_{ir,3} \in \mathbb{R}^{1 \times d}$ and $D_o^t \in \mathbb{R}^{H_f \times W_f}$.
With this attention matrix $D_{o}^t$, information can finally be extracted from the image $I \in \mathbb{R}^{H_f \times W_f \times d}$
by computing $E_{o}^t \in \mathbb{R}^d$:
\begin{equation}
    E_{o}^t = \sum_{h=1}^{H_f} \sum_{w=1}^{W_f} D_{o,h,w}^t I_{h, w}
\end{equation}
The output of this module is the \textbf{spatially-attended image feature} matrix $E^t \in \mathbb{R}^{O \times d}$ which 
contains all the $E_o^t$ vectors of the current reasoning step.

\subsection{Spatial Module}
\label{sect:gps}

The \texttt{Spatial} module is responsible for: 
(1) updating the previously extracted image data in $M^{t-1} \in \mathbb{R}^{O \times d}$ (section \ref{sect:update-spatial-mem}) with $E^t \in \mathbb{R}^{O \times d}$, 
(2) updating the score $s_{o}^t$ (section \ref{sect:update-score}) of each object based on the alignment between the updated extracted image data, $M^t_o$, and the query embedding $Q_e \in \mathbb{R}^{d}$, and (3) compute a new \texttt{SpatialMap} $S^t_o$ (section  \ref{sect:update-spatialmap}) based on the score $s_{o}^t$ and the initial \texttt{SpatialMap} $S^0_o$ of that object.
The inputs to the \texttt{Spatial} module are: (a) the extracted image data $E^t \in \mathbb{R}^{O \times d}$ from the \texttt{ImgReader} and (b) the query embedding $Q_e$.
The module will output an updated $M^t \in \mathbb{R}^{O \times d}$ matrix
and the updated \texttt{SpatialMap} tensor $S^t \in \mathbb{R}^{O \times H_f \times W_f}$.
Like in \texttt{ImgReader} we will add a subscript $o$ to indicate object specific data.
Before explaining in detail the operations in this module, we will start with explaining how the \texttt{SpatialMap} is built.

\subsubsection{SpatialMap}
\label{sect:gps-map}
\label{sect:gps-map-closer}

The \texttt{SpatialMap}, $S^t \in \mathbb{R}^{O \times H_f \times W_f}$, is a 
tensor that stores the location of the $O$ objects found by a RPN. 
At $t=0$, the \texttt{SpatialMap} is \textbf{initialized} by mapping 
every object $o$ onto a 2D grid $S^t_o \in \mathbb{R}^{H_f \times W_f}$,
assigning a weight of $1$ to every grid cell of this 2D map that falls inside $o$'s bounding box, and a lower weight, $x_s \in (0,1)$ outside $o$'s box.
We refer to this initial \texttt{SpatialMap} for an object $o$ as $S^0_o$. 
Further in this module we update this initial \texttt{SpatialMap} in every step $t$, which we refer to as $S^t_o$ (section \ref{sect:update-spatialmap}). The \texttt{SpatialMap} serves as a soft-attention 
over the image features $I$. 
Firstly, 
it indicates the location of the object $o$,
hence why grid cells that fall within an object box are given a weight of $1$. We also want the model to ``see'' the features in the grid cells that fall outside of the bounding box $o$, hence these regions receive a non-zero weight $x_s \in (0,1)$. For example, if we are looking for ``the man next to the tree'' and an entry in the \texttt{SpatialMap} indicates the location of a person, then the \texttt{ImgReader} module should look around this region for the tree. 
The computation of the \texttt{SpatialMap} can be seen in Figure \ref{fig:gps-idea}(b).

\subsubsection{Updating Spatial Memory}
\label{sect:update-spatial-mem}
To update the 
\textbf{memory} data $M_o^{t-1} \in \mathbb{R}^{d}$ with the current 
spatially-attended image features $E_o^t \in \mathbb{R}^{d}$ we merge them together in $U_{o}^t \in \mathbb{R}^{d}$:
\begin{equation}
    U_{o}^t = W_{sp,1} [E_{o}^t;M_{o}^{t-1}] + b,
    \label{eq:mem1}
\end{equation}
with $W_{sp,1} \in \mathbb{R}^{d \times 2d}$.
However, 
in order to enable data from previous reasoning steps to still be relevant in the current step, an attention distribution based on the similarity score between the embedding of the current focused words, $a^t \in \mathbb{R}^d$, and the embedding of the focused words $a^j \in \mathbb{R}^d$ from all the previous steps 
$j = 1, ..., t-1$, is calculated as:
\begin{equation}
    e^{t,j} = \texttt{softmax}(W_{sp,2} (a^t \odot a^{j}) + b),
\end{equation}
with $W_{sp,2} \in \mathbb{R}^{1\times d}$ and $\odot$ being the element-wise multiplication operator. Now, information from previous memory vectors are combined into $C_{o}^t \in \mathbb{R}^{d}$:
\begin{equation}
    C_{o}^t = \sum_{j=1}^{t-1} e^{t,j} M_{o}^j
\end{equation}
Finally, $U_{o}^t \in \mathbb{R}^{d}$ and $C_{o}^t \in \mathbb{R}^{d}$ are merged into the new 
memory vector $M_o^t \in \mathbb{R}^{d}$: 
\begin{equation}
\label{eq:uS}
    M_{o}^t = W_{sp,3}[U_{o}^t; C_{o}^t] + b,
\end{equation}
with $W_{sp,3} \in \mathbb{R}^{d \times 2d}$.

\subsubsection{Updating Scores}
\label{sect:update-score}
The scores are updated by calculating the alignment $V_{o}^t \in \mathbb{R}^{d}$ between the updated vector $M_{o}^t \in \mathbb{R}^{d}$ and the command $Q_e \in \mathbb{R}^{d}$.

\begin{equation}
    V_{o}^t = [W_{sp,4} M_{o}^t + b] \odot [W_{sp,5} Q_e + b],
\end{equation}
with $W_{sp,4} \in \mathbb{R}^{d \times d}$ and $W_{sp,5} \in \mathbb{R}^{d \times d}$.
Based on this alignment a score $s_{o}^t \in [0,1]$
is computed for each region. 
\begin{equation}
    s_{o}^t = \texttt{sigmoid}(W_{sp,6} V_{o}^t + b),
\end{equation}
with $W_{sp,6} \in \mathbb{R}^{1 \times d}$.
This \textbf{score} is used to determine the highest scoring object at the end of the reasoning process.

\subsubsection{Updating SpatialMap}
\label{sect:update-spatialmap}
Finally, to indicate the importance of an object $o$, we update the \texttt{SpatialMap} by multiplying the scores $s_{o}^t$ with the initial \texttt{SpatialMap} $S^0_o$ of $o$:

\begin{equation}
    S^t_o = s_{o}^t S_o^0
\end{equation}

If an object aligns well with the query $Q_e$ (thus $s_{o}^t$ is close to 1), this update will not substantially affect the initial \texttt{SpatialMap} $S_o^0$. Contrarily, low query-object match yields $s_{o}^t \approx 0$, resulting into $S^t_o$ close to zero everywhere and hence downplaying $o$. 



\subsection{Loss Function}
Training the network is seen as a multi-class classification problem where the network 
predicts which object region is the best region according to the query.
To this end, the cross-entropy loss is used which is defined in our case as:
\begin{equation}
\label{eq:loss}
   Loss = -\sum_{o=1}^{O} y_o log(s_{o}^T)
\end{equation}
where $s_{o}^T$ is the score of an object $o$ at the last reasoning step $T$ and where $y_o$ is 1 if the object $o$ 
is the referred object and 0 otherwise.

\section{Datasets}


In the experiments below we evaluate our model on two datasets: Talk2Car \cite{deruyttere2019talk2car} and RefCOCO \cite{yu2016modeling}.

\subsection{Talk2Car}
The Talk2Car \cite{deruyttere2019talk2car} dataset contains images from the nuScenes dataset \cite{nuScenes} that are annotated with natural language commands for self-driving cars, bounding boxes of scene objects, and the bounding box of the object that is referred to in a command. 
In total it contains 11,959 commands that belong to 9,217 images, which are either taken in Singapore or Boston during different weather (sun or rain) and time conditions (night or day).
On average, a command and an image each contain respectively around 11 words and 11 objects from 23 categories.
This dataset is selected because of its complex natural language commands that constrain -- through modifying language expressions -- the object to be found in the scene demanding reasoning over the object in the scene and words in the command. 
Train, validation and test sets contain respectively 8,349,  
1,163 
and 2,447 
commands.
In addition, the dataset consists of several smaller test sets, each of which evaluate specific challenging settings. 
A first sub-test set assesses the ability of a model to recognise distant referred objects.
The second and third sub-test sets allow evaluating how well a model can cope with short and long commands respectively. 
The final sub-test set assesses how the model copes with ambiguity. In our case ambiguity refers to having multiple objects of the referred class in the visual scene.
An example of the Talk2Car dataset can be seen in Figure \ref{fig:example-fig}.

\subsection{RefCOCO}
The RefCOCO dataset \cite{yu2016modeling} is based on the MS-COCO dataset \cite{MSCOCO} and contains 142,209 referring expressions for 50,000 objects and on average these expressions contain 3.61 words. The dataset is split into 40,000 objects for training, 5,000 for validation and 5,000 for testing. The latter is split in two test sets: ``TestA'' and ``TestB''.
The first test set, ``TestA'' consists of images that contain multiple people. The other images are put into ``TestB''. As \cite{deng2018visual,yu2016modeling,yu2017joint}, we will use the provided object regions of the MSCOCO dataset to train and evaluate our model.

\section{Experimental Setup}
In our experiments 
we evaluate the proposed \texttt{MSRR} and 
five different strong baselines (section \ref{sect:baselines}) on three different measures (section \ref{sect:metrics}). 
For model parameters, optimizers, initialization of certain vectors and qualitative examples we refer to the supplementary material.
\subsection{Baselines}
\label{sect:baselines}

As baselines we 
use \texttt{SCRC} \cite{hu2016natural}, an adapted \texttt{MAC} \cite{Hudson2018} for Visual Grounding \cite{deruyttere2019talk2car}, \texttt{STACK} \cite{hu2018explainable} and \texttt{A-ATT} \cite{deng2018visual}.
The difference between \texttt{MSRR} and \texttt{MAC} with \texttt{STACK} is that the latter two models, although both being multi-step reasoning models, do not reason over the regions themselves and do not use spatial information.
The difference between \texttt{MSRR} and \texttt{A-ATT} is the 
strategy employed to encode the spatial information. The latter extracts local information from an object and embeds it with the coordinates of said object.
We 
also include an ablated model called \texttt{MaskObjs} which uses a \texttt{SpatialMap} where only the object regions receive a weight of 1, while the grid cells that fall outside receive a 
weight of 0. This \texttt{SpatialMap} is then multiplied with the extracted image features to limit the search space to only the object regions. These transformed image features are passed to \texttt{MAC} to reason with. 

\subsection{Measures}
\label{sect:metrics}

All models are evaluated with three measures. 
The first measure is the overall \textit{accuracy $IoU_{0.5}$}, defined as the percentage of predicted regions that have an Intersection over Union (IoU) or overlap, with the ground truth regions of over $0.5$.
The second measure 
is \textit{inference time}, and the third one
is the \textit{number of parameters} of each model.

\section{Results}
In this section we will discuss the results for the Talk2Car dataset and the RefCOCO dataset.

\begin{table*}[t]
\small
\begin{center}
\resizebox{0.6\columnwidth}{!}{%
\begin{tabular}{|l|c|c|c|}
\hline
Method & $IoU_{0.5}$ (\%) & Inference Time (ms) & Params (M) \\
\hline
MAC \cite{Hudson2018}
& 50.51 & 51
& 41.59 \\
STACK \cite{hu2018explainable} & 33.71 & 52 & 35.2\\
SCRC (Top-32) \cite{hu2016natural} & 43.80 & 208 & 52.47 \\
A-ATT (Top-16) \cite{deng2018visual} & 45.12 & 180 & 160.31\\ 
\hline
MaskObjs (Top-16) & 54.31 & 270.5 & 62.25\\
MSRR (top-8) & 56.85 & 224.7 & 62.25 \\
MSRR (top-16) &\textbf{60.04} & 270.5 & 62.25 \\
MSRR (top-32) & 59.91 & 359.7 & 62.39 \\
MSRR (top-64) & 58.85 & 576.2 & 62.39 \\
\hline
\end{tabular}
}
\end{center}
\caption{Performance ($IoU_{0.5}$), inference time (evaluated on a TITAN XP) and number of parameters of the different models. All models that use object regions have been evaluated with the top-$k$ ($k=8,16,32,64$) scoring regions. In the table we only display the best $k$-value for \texttt{SCRC}, \texttt{MaskObjs} and \texttt{A-ATT} for brevity. Results are for Talk2Car dataset. For our models we show results that are averaged over three runs.}
\label{tab:experiments}
\end{table*}

\begin{figure*}[t]
\begin{center}
\includestandalone[width=1\linewidth]{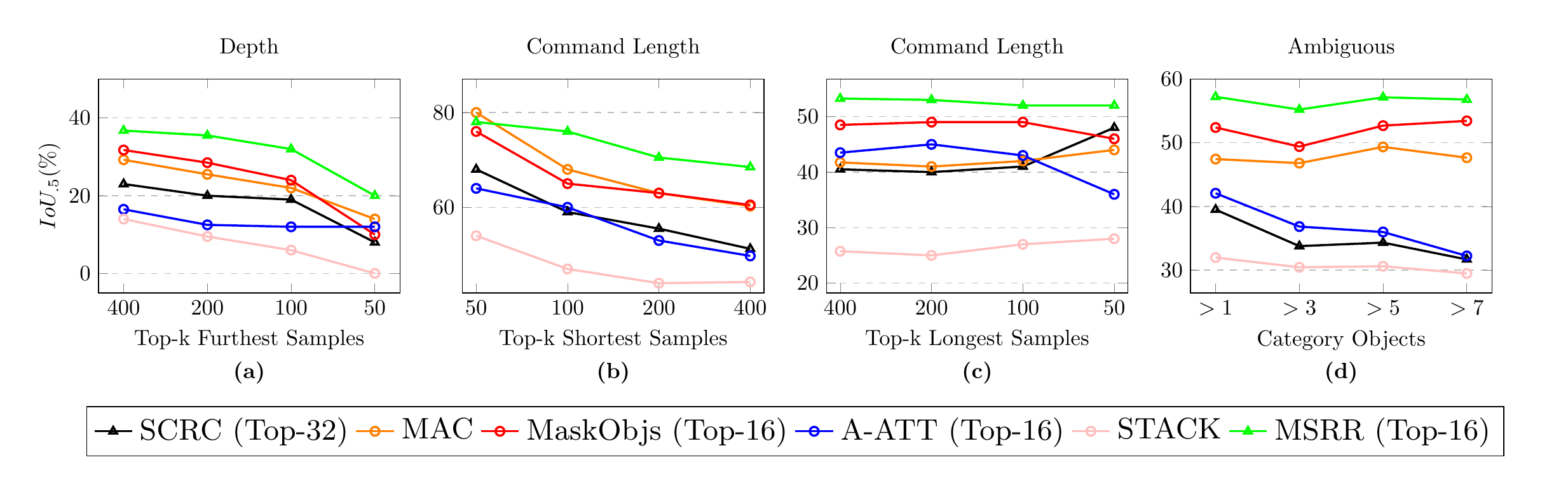}
\caption{
Results of the best models on the four different sub-test sets in the Talk2Car dataset. Each plot starts with the easy cases on the left side and moves to harder cases on the right. For plot (a), (b), (c) the x-axis represents the number of samples used in the test case. For plot (d) the x-axis represents the number of objects that have the same class as the referred object in a certain image.} 
\label{fig:subset_tests}
\end{center}
\end{figure*}

\subsection{Talk2Car}
The results of our \texttt{MSRR} compared to our 
four baselines and the state-of-the-art (\texttt{MAC}) on the Talk2Car test set are shown in Table \ref{tab:experiments}.
We observe that \texttt{MSRR} clearly outperforms 
the rest of models for any top-$k$ 
number of regions. 
Even though this top-$k$ confidence selection mechanism of bounding boxes is 
simple,
it provides a substantial improvement.
The best \texttt{MSRR} model further improves the state-of-the-art baseline (\texttt{MAC}) nearly 10\% in terms of IoU. Our model is however five times slower than \texttt{MAC} and has 20M additional parameters.
We argue that the difference in both accuracy and inference times come from the fact that our model reasons over each object in the image 
while \texttt{MAC} reasons solely over the whole image.
Results on the four difficult sub-test sets (Figure \ref{fig:subset_tests}) 
show that our best model substantially and consistently outperforms the other models.

\subsection{RefCOCO}
\label{sect:refcoco}
The natural context of our proposed model are complex VG tasks such as Talk2Car, which require multiple reasoning steps to correctly locate the referred object in the image. However, for completeness and as a robustness test we additionally evaluate 
our model in the RefCOCO task. In particular, the average \textit{query sentence length} in RefCOCO is 3.61 while in Talk2Car is 11. Results align with this analysis of the task's complexity / difficulty. That is, the models generally perform better with a small number of reasoning steps (two) than with a large one (ten). In particular, \texttt{MSRR} obtains an $IoU_{0.5}$ of (77.73, 76.31) = ('test A', 'test B') with 2 reasoning steps and $IoU_{0.5}$ of (75.2, 74.25) with 10 steps. \texttt{MAC} obtains an $IoU_{0.5}$ of (51.23, 38) = ('test A', 'test B') with 2 reasoning steps and $IoU_{0.5}$ of (52.01, 37.17) with 10 steps.
This contrasts with Talk2Car, where \texttt{MSRR} (top-16) tends to rather increase performance (and stability) with the number of reasoning steps (Figure \ref{fig:multistep_experiments}(a)).
These results indicate that RefCOCO's task is rather simple and does not seem to require multiple step reasoning. Overall, \texttt{MSRR} shows strong results in RefCOCO, yet not better than 
\texttt{A-ATT} which attains an $IoU_{0.5}$ of (81.67, 79.96) = ('test A', 'test B'). Contrarily, in Talk2Car \texttt{MSRR} clearly outperforms \texttt{A-ATT}. 
This is not surprising given that 
\texttt{MSRR} is meant for complex tasks that require multi-step reasoning. 


\subsection{Empirically Attainable Bayes Performance on Talk2Car}
We perform two tests on Talk2Car to determine the performance bottlenecks of our model.
There are two main possible sources: either the predicted object regions are 
weak or the model itself can be improved.
First, we asses the quality of the \textbf{predicted object regions (boxes)} without using the model. We consider a prediction as correct if there exist one predicted box that overlaps more than 50\% with the ground truth box answer. 
We find that with the top-64 regions a performance of 93.58\% $IoU_{0.5}$ is attainable.
This implies that for $\approx$ 6.5\% of the test set there is no predicted bounding box for the referred object. 
When considering only the top-8, top-16 and top-32 object regions 
the attainable $IoU_{0.5}$ are 79.85\%, 88.14\% and 91.74\%, respectively. 
The high attainable performance suggests that the predicted regions are of good quality. 

Second, we train and evaluate our
model with the \textbf{ground truth object regions} 
and we find that the model attains 68\% $IoU_{0.5}$ accuracy.
This suggest that the model still has room for improvement, yet model weakness may not be the only factor keeping us from a perfect score (100\%). Intrinsic predictability limitations of the task such as annotation errors, noise may also account for a proportion of the missing 32\% (= 100 - 68).
Furthermore, the comparison between the 68\% $IoU_{0.5}$ attained with ground truth boxes and the 60.29\% with predicted boxes (Table \ref{tab:experiments}) reveals that our model misses $\approx$ 8\% accuracy due to the weakness of the predicted boxes.


\subsection{Ablation Study}

We subsequently describe a number of ablation experiments performed in the Talk2Car dataset, which reveal the contribution of different elements and modules of our model (\texttt{MSRR}) onto its overall performance.

\subsubsection{Spatial Module} 
The baseline \texttt{MAC} \cite{Hudson2018} constitutes a functional ablation of our model. More concretely, \texttt{MAC} reasons over the whole image and predicts the coordinates of the object that matches with the referring expression, while our model reasons over the image but also over the extracted objects to select one of 
them as the answer. 
Thus, \texttt{MAC} can be seen as an ablated version of our model, with no \texttt{Spatial} module. The improvement of \texttt{MSRR} over \texttt{MAC} (Table \ref{tab:experiments}) illustrates thus the importance of the \texttt{Spatial} module. 


\subsubsection{Soft Spatial Attention} 
We recall that \texttt{SpatialMap} 
assigns a value $x_s \in [0,1]$ to grid cells that do not belong to a region (section \ref{sect:gps-map}). Figure \ref{fig:multistep_experiments}(b) shows the influence of this value, where three important cases can be distinguished.

\begin{itemize}
    \item $\boldsymbol{x_s=0}$ \textbf{(\texttt{MaskObjs})}: 
\texttt{SpatialMap} $S_o \in \mathbb{R}^{H_f \times W_f}$ of an object $o$ contains only ones at the grid cells within the object and zeros everywhere outside. 
Hence, the product of the image features $I$ with the \texttt{SpatialMap} $S_o$ (eq. \ref{eq:mult-spatial}) keeps only the features of the object while setting the features outside of it to 0. This can be seen as a hard-attention mechanism. 
The only information that is available to the model is the local information of the object together with its spatial location.
The model can thus not extract anything sensible from the grid cells around the object anymore as they are all zero.
However, 
the model can still get an accuracy of 54.31\% 
(blue value in figure \ref{fig:multistep_experiments}(b)),  
outperforming the previous state of the art.

\item $\boldsymbol{x_s=1}$:  
The \texttt{SpatialMap} $S_o \in \mathbb{R}^{H_f \times W_f}$ is filled with ones, which results in having the same \texttt{SpatialMap} for every object. 
When we multiply $S_o$ with the image features $I$, we get the original image features back. The model can thus not reason over the objects anymore and must guess which one of the top-k objects is the best one. The resulting accuracy of 
this model is: 7.76\%  
(red value in figure \ref{fig:multistep_experiments}(b)). To prove that the case $x_s=1$ 
is the same as \textbf{randomly choosing an object}, we implemented 
a model that randomly selects one of the top-16 objects as answer. The resulting $IoU_{0.5}$ accuracy when averaged over 1,000 runs is 7.77\%. 

\item $\boldsymbol{0 < x_s < 1}$ \textbf{(\texttt{MSRR}): } 
When $x_s \in (0,1)$, the accuracy of the model improves drastically compared to $x_s=0$ and $x_s=1$.
The highest accuracy (60.29\%) is attained when $x_s=0.5$ 
(green value in figure \ref{fig:multistep_experiments}(b)).
Hence, both allowing the model to look at features around the objects 
and giving these features less importance than the object features seem crucial for this task.

\end{itemize}

\subsubsection{Number of Reasoning Steps Ablation}
\begin{figure}
\begin{center}
\begin{minipage}{0.45\textwidth}
        \centering
\includestandalone[width=\linewidth]{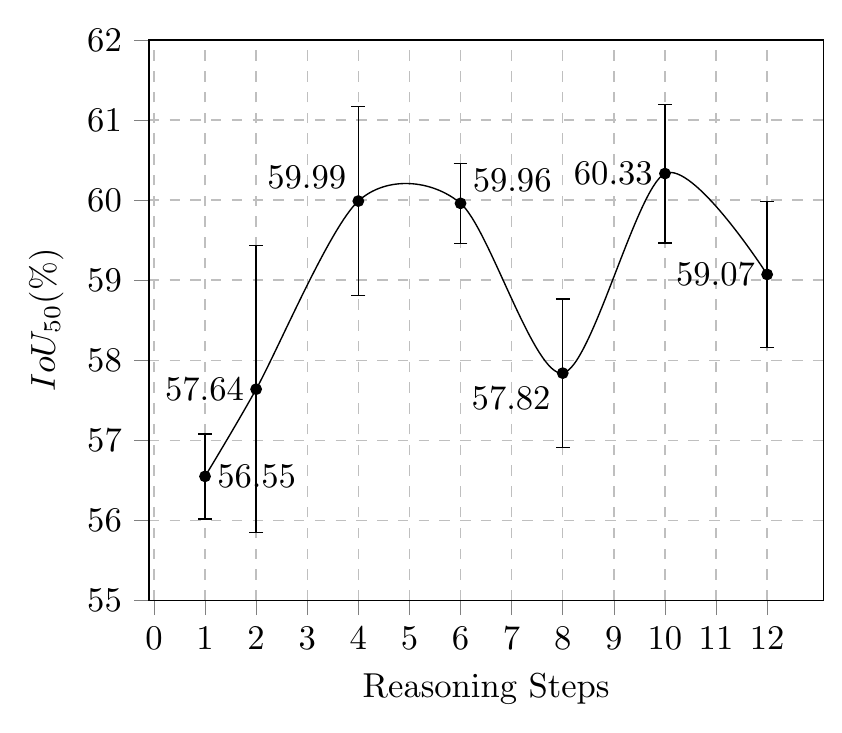}
(a)
\end{minipage}
\begin{minipage}{0.45\textwidth}
        \centering
\includestandalone[width=\linewidth]{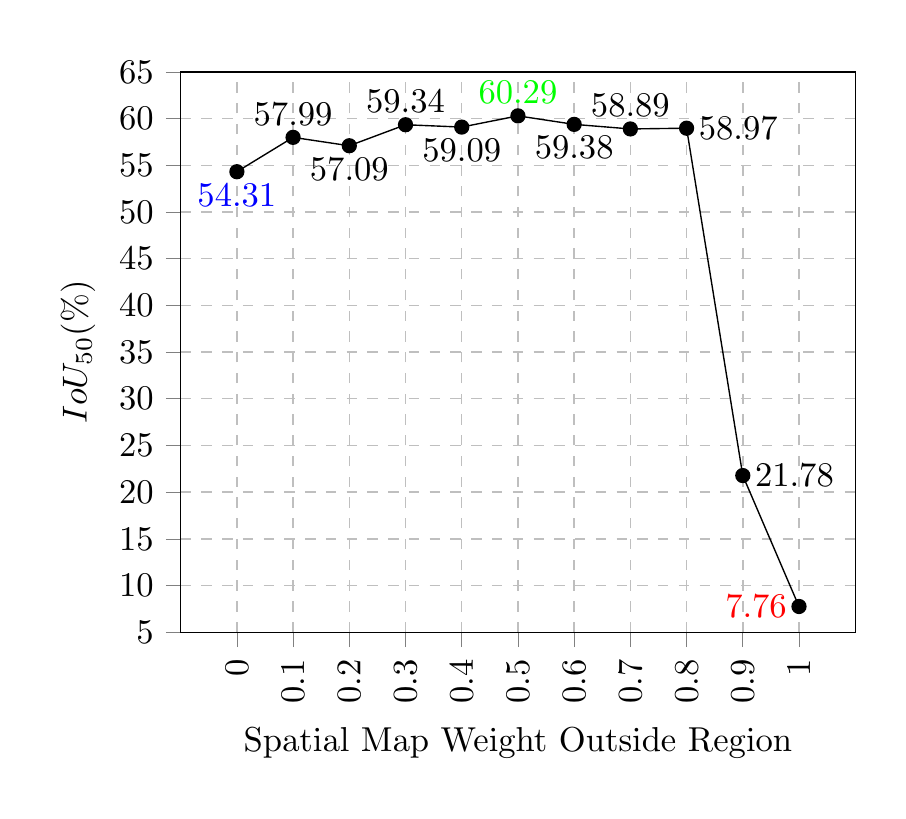}
(b)
\end{minipage}

\caption{
(a) The influence of different reasoning steps on the $IoU_{0.5}$ of the \texttt{MSRR} model is shown for the Talk2Car validation set. We also include the variance of three different runs. (b) The influence on the $IoU_{0.5}$ of the \texttt{MSRR} model with 10 reasoning steps when changing the $x_s$ value used in the \texttt{SpatialMap} for grid cells that do not belong to an object. 
} 
\label{fig:multistep_experiments}
\end{center}
\end{figure}

The influence of the number of reasoning steps on the $IoU_{0.5}$ on the validation set for a \texttt{MSRR} (top-16) model with $x_s = 0.5$ 
is shown in figure \ref{fig:multistep_experiments}(a). 
We notice that a single reasoning step gives the lowest $IoU_{0.5}$ accuracy (56.55\%),
and that by increasing the number of reasoning steps we can gain around 4\% $IoU_{0.5}$ by using 10 reasoning steps.
We also observe that for a certain number of reasoning steps (e.g., 2), there is a high variance on the model's performance across runs. 
Ten reasoning steps 
provided the highest average accuracy out of three runs and hence we use this value for the test set.


         
        

\section{Conclusions and Future Work}

In this paper we have introduced \texttt{MSRR}, a multimodal and multi-step reasoning model for visual grounding tasks. 
The main evaluation of the model is carried out in the Talk2Car dataset which is composed of images of city environments taken from the viewpoint of a car and accompanied by commands that passengers give to the car. 
The proposed model that jointly reasons over the words of the command and the detected objects in the image
outperforms state-of-the-art models by a large margin in this task.
We further validate our model in the RefCOCO dataset, which is markedly simpler than Talk2car. 
Although our model fares well in RefCOCO, we show that performing multi-step reasoning is rather detrimental in this task. We highlight that the natural context of \texttt{MSRR} are rather complex visual grounding tasks that require multi-step reasoning, such as Talk2Car.   
Additionally, having a separate reasoning process for each object, thanks to the \texttt{SpatialMap}, is certainly beneficial in environments like self-driving cars as it allows the reasoning process to remain transparent but also indicates when the model is hesitant in certain situations based on the scores of certain objects. 

Furthermore, we perform diverse ablation studies to elucidate the contribution of the different modules on \texttt{MSRR}'s performance, including the ablation of the spatial module, the number of reasoning steps and the weight of regions outside the objects. 

Overall, the proposed model naturally suggests several opportunities for future improvements not explored in this paper yet. 
For instance, one may take into account object classes recognised in the image and their probability distributions (cf. as done in linking text entities with knowledge base entities in \cite{DBLP:conf/acl/LeT19}), or integrating intelligent selection mechanisms that give priority to the processing of certain words or objects based on prior knowledge.

\section{Acknowledgments}{
This project is sponsored by the MACCHINA project from the KU Leuven  with grant number C14/18/065 and also by the European Research Council Advanced Grant H2020-ERC-2017-ADG 788506.
We would like to thank Nvidia for granting us two TITAN Xp GPUs.
}

\newpage
{\small
\bibliographystyle{splncs04}
\bibliography{main}

\begin{thebibliography}{10}
\providecommand{\url}[1]{\texttt{#1}}
\providecommand{\urlprefix}{URL }
\providecommand{\doi}[1]{https://doi.org/#1}

\bibitem{nuScenes}
Caesar, H., Bankiti, V., Lang, A.H., Vora, S., Liong, V.E., Xu, Q., Krishnan,
  A., Pan, Y., Baldan, G., Beijbom, O.: nuscenes: A multimodal dataset for
  autonomous driving. arXiv preprint arXiv:1903.11027  (2019)

\bibitem{DBLP:conf/iccv/ChenG17}
Chen, X., Gupta, A.: Spatial memory for context reasoning in object detection.
  In: {IEEE} International Conference on Computer Vision, {ICCV} 2017, Venice,
  Italy, October 22-29, 2017. pp. 4106--4116 (2017).
  \doi{10.1109/ICCV.2017.440},
  \url{http://doi.ieeecomputersociety.org/10.1109/ICCV.2017.440}

\bibitem{deng2018visual}
Deng, C., Wu, Q., Wu, Q., Hu, F., Lyu, F., Tan, M.: Visual grounding via
  accumulated attention. In: Proceedings of the IEEE Conference on Computer
  Vision and Pattern Recognition. pp. 7746--7755 (2018)

\bibitem{deng2009imagenet}
Deng, J., Dong, W., Socher, R., Li, L.J., Li, K., Fei-Fei, L.: Imagenet: A
  large-scale hierarchical image database. In: 2009 IEEE conference on computer
  vision and pattern recognition. pp. 248--255. Ieee (2009)

\bibitem{deruyttere2019talk2car}
Deruyttere, T., Vandenhende, S., Grujicic, D., Van~Gool, L., Moens, M.F.:
  Talk2car: Taking control of your self-driving car. In: Proceedings of the
  2019 Conference on Empirical Methods in Natural Language Processing and the
  9th International Joint Conference on Natural Language Processing
  (EMNLP-IJCNLP). pp. 2088--2098 (2019)

\bibitem{FukuiPYRDR16}
Fukui, A., Park, D.H., Yang, D., Rohrbach, A., Darrell, T., Rohrbach, M.:
  Multimodal compact bilinear pooling for visual question answering and visual
  grounding. CoRR  \textbf{abs/1606.01847} (2016),
  \url{http://arxiv.org/abs/1606.01847}

\bibitem{DBLP:conf/cvpr/GuptaDLSM17}
Gupta, S., Davidson, J., Levine, S., Sukthankar, R., Malik, J.: Cognitive
  mapping and planning for visual navigation. In: 2017 {IEEE} Conference on
  Computer Vision and Pattern Recognition, {CVPR} 2017, Honolulu, HI, USA, July
  21-26, 2017. pp. 7272--7281 (2017). \doi{10.1109/CVPR.2017.769},
  \url{https://doi.org/10.1109/CVPR.2017.769}

\bibitem{ResNet}
He, K., Zhang, X., Ren, S., Sun, J.: Deep residual learning for image
  recognition. CoRR  \textbf{abs/1512.03385} (2015),
  \url{http://arxiv.org/abs/1512.03385}

\bibitem{Henriques_2018_CVPR}
Henriques, J.F., Vedaldi, A.: Mapnet: An allocentric spatial memory for mapping
  environments. In: The IEEE Conference on Computer Vision and Pattern
  Recognition (CVPR) (June 2018)

\bibitem{hu2018explainable}
Hu, R., Andreas, J., Darrell, T., Saenko, K.: Explainable neural computation
  via stack neural module networks. In: Proceedings of the European Conference
  on Computer Vision (ECCV). pp. 53--69 (2018)

\bibitem{hu2016natural}
Hu, R., Xu, H., Rohrbach, M., Feng, J., Saenko, K., Darrell, T.: Natural
  language object retrieval. In: Proceedings of the IEEE Conference on Computer
  Vision and Pattern Recognition. pp. 4555--4564 (2016)

\bibitem{Hudson2018}
Hudson, D.A., Manning, C.D.: Compositional attention networks for machine
  reasoning. CoRR  \textbf{abs/1803.03067} (2018),
  \url{http://arxiv.org/abs/1803.03067}

\bibitem{Johnson2017}
Johnson, J., Hariharan, B., Maaten, L.V.D., Hoffman, J., Fei-Fei, L., Zitnick,
  C.L., Girshick, R.: {Inferring and Executing Programs for Visual Reasoning}.
  In: Proceedings of the IEEE International Conference on Computer Vision. vol.
  2017-Octob, pp. 3008--3017 (2017). \doi{10.1109/ICCV.2017.325}

\bibitem{kafle2016answer}
Kafle, K., Kanan, C.: Answer-type prediction for visual question answering. In:
  Proceedings of the IEEE Conference on Computer Vision and Pattern
  Recognition. pp. 4976--4984 (2016)

\bibitem{karpathy2014deep}
Karpathy, A., Joulin, A., Fei-Fei, L.F.: Deep fragment embeddings for
  bidirectional image sentence mapping. In: Advances in neural information
  processing systems. pp. 1889--1897 (2014)

\bibitem{kingma2014adam}
Kingma, D.P., Ba, J.: Adam: A method for stochastic optimization. arXiv
  preprint arXiv:1412.6980  (2014)

\bibitem{DBLP:conf/acl/LeT19}
Le, P., Titov, I.: Boosting entity linking performance by leveraging unlabeled
  documents. In: Proceedings of the 57th Conference of the Association for
  Computational Linguistics, {ACL} 2019, Florence, Italy, July 28- August 2,
  2019, Volume 1: Long Papers. pp. 1935--1945 (2019),
  \url{https://www.aclweb.org/anthology/P19-1187/}

\bibitem{MSCOCO}
Lin, T., Maire, M., Belongie, S.J., Bourdev, L.D., Girshick, R.B., Hays, J.,
  Perona, P., Ramanan, D., Doll{\'{a}}r, P., Zitnick, C.L.: Microsoft {COCO:}
  common objects in context. CoRR  \textbf{abs/1405.0312} (2014),
  \url{http://arxiv.org/abs/1405.0312}

\bibitem{liu2016ssd}
Liu, W., Anguelov, D., Erhan, D., Szegedy, C., Reed, S., Fu, C.Y., Berg, A.C.:
  Ssd: Single shot multibox detector. In: European conference on computer
  vision. pp. 21--37. Springer (2016)

\bibitem{liu2019improving}
Liu, X., Wang, Z., Shao, J., Wang, X., Li, H.: Improving referring expression
  grounding with cross-modal attention-guided erasing. In: Proceedings of the
  IEEE Conference on Computer Vision and Pattern Recognition. pp. 1950--1959
  (2019)

\bibitem{mao2016generation}
Mao, J., Huang, J., Toshev, A., Camburu, O., Yuille, A.L., Murphy, K.:
  Generation and comprehension of unambiguous object descriptions. In:
  Proceedings of the IEEE conference on computer vision and pattern
  recognition. pp. 11--20 (2016)

\bibitem{DBLP:conf/iclr/ParisottoS18}
Parisotto, E., Salakhutdinov, R.: Neural map: Structured memory for deep
  reinforcement learning. In: 6th International Conference on Learning
  Representations, {ICLR} 2018, Vancouver, BC, Canada, April 30 - May 3, 2018,
  Conference Track Proceedings (2018),
  \url{https://openreview.net/forum?id=Bk9zbyZCZ}

\bibitem{qian1999momentum}
Qian, N.: On the momentum term in gradient descent learning algorithms. Neural
  networks  \textbf{12}(1),  145--151 (1999)

\bibitem{Suarez2018}
Suarez, J., Johnson, J., Li, F.F.: {DDRprog: A CLEVR Differentiable Dynamic
  Reasoning Programmer}  (2018), \url{http://arxiv.org/abs/1803.11361}

\bibitem{yu2018mattnet}
Yu, L., Lin, Z., Shen, X., Yang, J., Lu, X., Bansal, M., Berg, T.L.: Mattnet:
  Modular attention network for referring expression comprehension. In: CVPR
  (2018)

\bibitem{yu2016modeling}
Yu, L., Poirson, P., Yang, S., Berg, A.C., Berg, T.L.: Modeling context in
  referring expressions. In: European Conference on Computer Vision. pp.
  69--85. Springer (2016)

\bibitem{yu2017joint}
Yu, L., Tan, H., Bansal, M., Berg, T.L.: A joint speaker-listener-reinforcer
  model for referring expressions. In: Proceedings of the IEEE Conference on
  Computer Vision and Pattern Recognition. pp. 7282--7290 (2017)

\bibitem{zhang2015accelerating}
Zhang, X., Zou, J., He, K., Sun, J.: Accelerating very deep convolutional
  networks for classification and detection. IEEE transactions on pattern
  analysis and machine intelligence  \textbf{38}(10),  1943--1955 (2015)

\bibitem{ZhouTSSF15}
Zhou, B., Tian, Y., Sukhbaatar, S., Szlam, A., Fergus, R.: Simple baseline for
  visual question answering. CoRR  \textbf{abs/1512.02167} (2015),
  \url{http://arxiv.org/abs/1512.02167}

\bibitem{zhou2019objects}
Zhou, X., Wang, D., Kr{\"a}henb{\"u}hl, P.: Objects as points. arXiv preprint
  arXiv:1904.07850  (2019)

\end{thebibliography}
}
\clearpage
\section{Supplementary Material}
\subsection{Model Parameters}
\subsubsection{Talk2Car}
To extract the $O$ object objects from each image of the dataset, we pre-train a CenterNet model \cite{zhou2019objects} on the images of the Talk2Car training set. 
The models that use extracted objects will use this network to perform this task.
Guided by \cite{deruyttere2019talk2car} we have set the amount of reasoning steps of \texttt{MAC} to $10$.
Based on a development set we found that for  \texttt{MSRR} and \texttt{MaskObjs} $10$ reasoning steps works the best too (see figure \ref{fig:multistep_experiments}(a)).
In a same way we have set the learning rate for these three models to $0.0001$ by using the Adam optimizer \cite{kingma2014adam}. The $d$-parameter in all the models is set to 512.
Additionally, during training of the \texttt{MSRR} model we apply gradient clipping by the global norm if the global norm is higher than $5$.
The \texttt{SCRC} model was initialized as described in \cite{deruyttere2019talk2car} except that we did not use regions extracted with a SSD-512 network\cite{liu2016ssd} but with CenterNet.
For the \texttt{A-ATT} model, the amount of reasoning steps have been set to 4 according to \cite{deng2018visual} and empirical tests on the development set.
The learning rate for this model has empirically been set to $0.001$ and the SGD optimizer with momentum \cite{qian1999momentum} has been used as was done by \cite{deng2018visual}.
In this paper, to vary the amount of regions, we use a simple selection criteria by simply taking the top-$k$ confident regions based on the confidence score of the regions extracted by CenterNet.

\subsubsection{RefCOCO}
For our experiments on the RefCOCO dataset we follow the approach of \cite{deng2018visual,MSCOCO}: we train and evaluate our \texttt{MSRR} model with the ground truth regions. We, however, found by using a development set, that using less reasoning steps (2 vs 10) improved the accuracy of our model (section \ref{sect:refcoco}).

\subsection{Image Size}
\subsubsection{Talk2Car}
The original size of an image in the Talk2Car dataset is 1600 (width) by 900 (height). To extract features for the models, we resize the image to 512 by 512 and we cut off a ResNet-101 \cite{ResNet} model pre-trained on ImageNet \cite{deng2009imagenet} at the fourth layer as is done by \cite{Hudson2018,hu2018explainable}.
The extracted image features have the shape [$H_f \times W_f \times d$] with $H_f$ = 32, $W_f$ = 32 and $d$ = 1024. 
We wish to mention that for the \texttt{A-ATT} model we found that we had to resize the images to 224 by 224 to have a good accuracy with the model. This leads to image features with $H_f$ = 14, $W_f$ = 14 and $d$ = 1024. 
We also tested the same resolution with other models but we did not see any increase in accuracy.

\subsubsection{RefCOCO}
The images in the RefCOCO dataset have varying sizes and were all resized to $448$ by $448$.
To extract the features we used a pre-trained VGG-16\cite{zhang2015accelerating} on ImageNet\cite{deng2009imagenet}. This network was cut off at the last convolutional layer as is done by \cite{deng2018visual}. The extracted image features have the shape [$H_f \times W_f \times d$] with $H_f$ = 14, $W_f$ = 14 and $d$ = 512. 

\subsection{Qualitative Results}
\begin{figure*}
\begin{center}
  \begin{tabular}{ c c }
    \textbf{MSRR} & \textbf{MAC} \\
    \begin{minipage}{0.45\textwidth}
      \includegraphics[width=\textwidth]{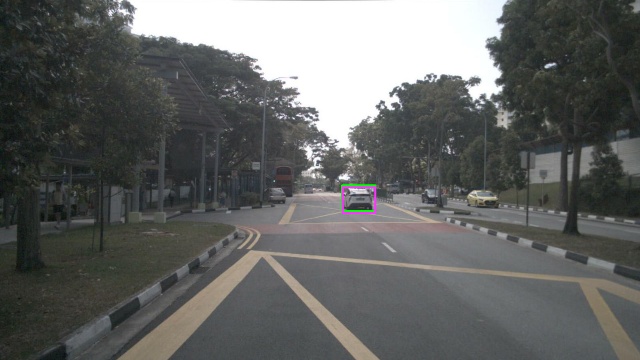}
    \end{minipage}
    & 
    \begin{minipage}{0.45\textwidth}
      \includegraphics[width=\textwidth]{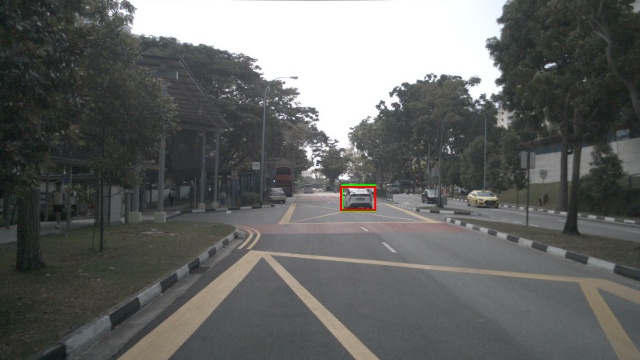}
    \end{minipage} \\
    \multicolumn{2}{C{12cm}}{    \textbf{Command}: Change lanes and get behind \textbf{the white car }
}
 \\
    \begin{minipage}{0.45\textwidth}
  \includegraphics[width=\textwidth]{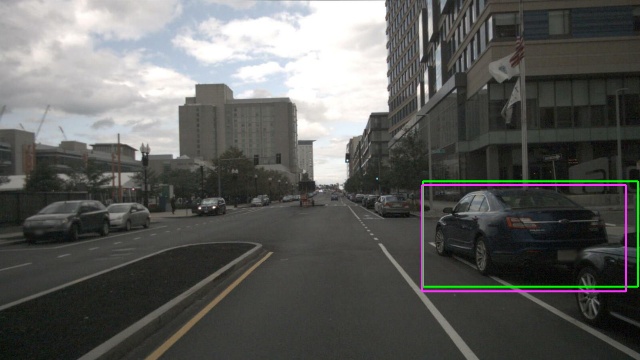}
    \end{minipage}
    & 
    \begin{minipage}{0.45\textwidth}
\includegraphics[width=\textwidth]{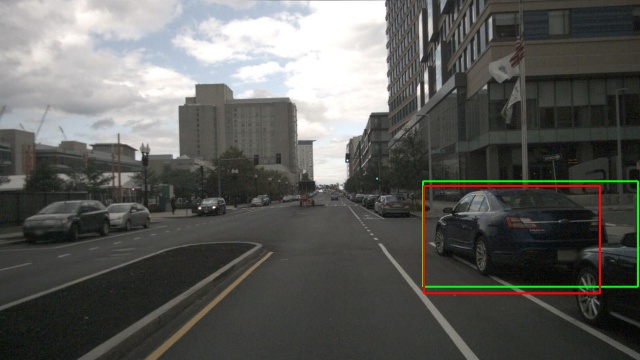}
    \end{minipage} \\
    \multicolumn{2}{C{12cm}}{    \textbf{Command}: Take a right after \textbf{the car on our right}}.
\\
    \begin{minipage}{0.45\textwidth}
  \includegraphics[width=\textwidth]{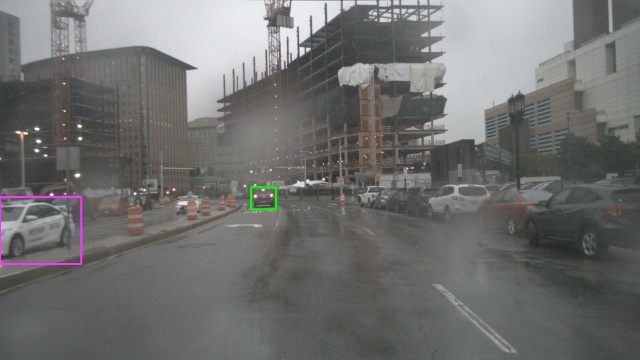}
    \end{minipage}
    & 
    \begin{minipage}{0.45\textwidth}
\includegraphics[width=\textwidth]{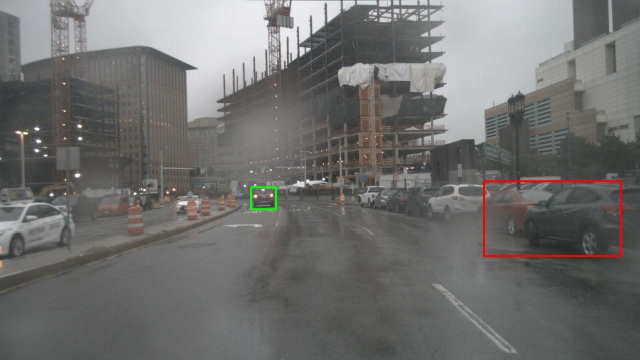}
    \end{minipage} \\
    \multicolumn{2}{C{12cm}}{    \textbf{Command}: Make the left turn like \textbf{that car in front of us}.
} \\

\begin{minipage}{0.45\textwidth}
  \includegraphics[width=\textwidth]{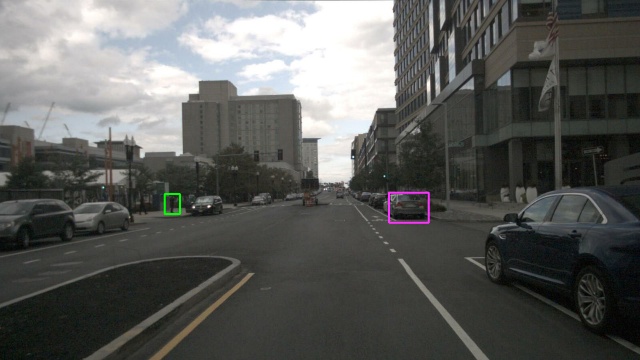}
    \end{minipage}
    & 
    \begin{minipage}{0.45\textwidth}
\includegraphics[width=\textwidth]{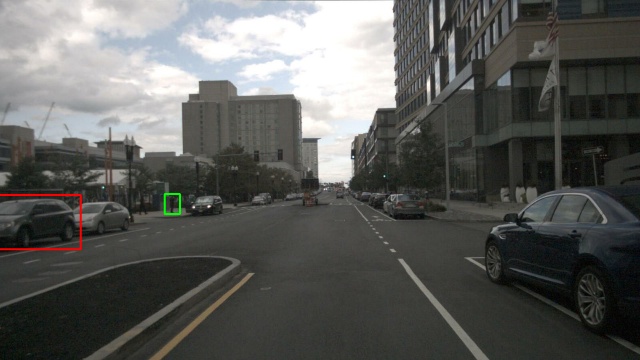}
    \end{minipage} \\
    \multicolumn{2}{C{12cm}}{    \textbf{Command}: I think that is \textbf{Jim over there next to that parked black car on the other side}. Stop when parallel to him.
}
\end{tabular}
\end{center}
\caption{Examples from the test set showing two successful groundings by the \texttt{MSRR} (left) and \texttt{MAC} (right) in the first two images and two failures in the two final images. For each image, the command is given together with the referred object indicated in bold in the text. In the images themselves, the green bounding box indicates the ground truth, purple indicates the output of the \texttt{MSRR} and red the output bounding box of \texttt{MAC}.}
\label{fig:success-failure-cases}
\end{figure*}

\begin{figure*}[h!]
  \begin{center}
  \begin{tabular}{ c c }
    \textbf{MSRR} & \textbf{MAC} \\
    \begin{minipage}{0.45\textwidth}
      \includegraphics[width=\textwidth]{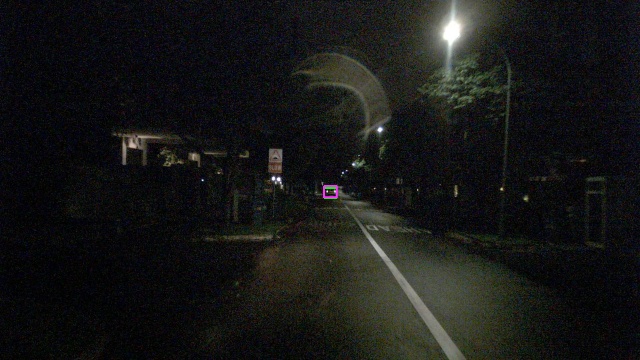}
    \end{minipage}
    & 
    \begin{minipage}{0.45\textwidth}
      \includegraphics[width=\textwidth]{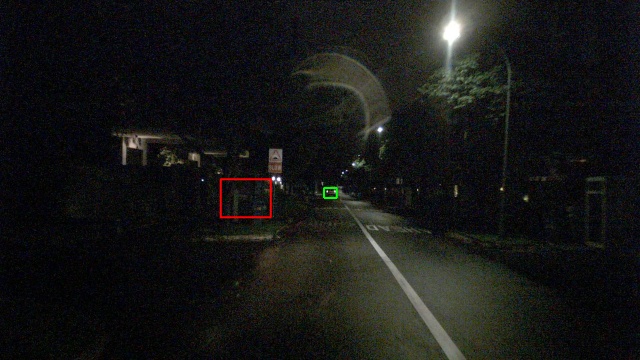}
    \end{minipage} \\
    \multicolumn{2}{c}{    \textbf{Command}: \textbf{That car in front of us}! That is my brother! Follow him until he stops.
} \\
    \begin{minipage}{0.45\textwidth}
   \includegraphics[width=\textwidth]{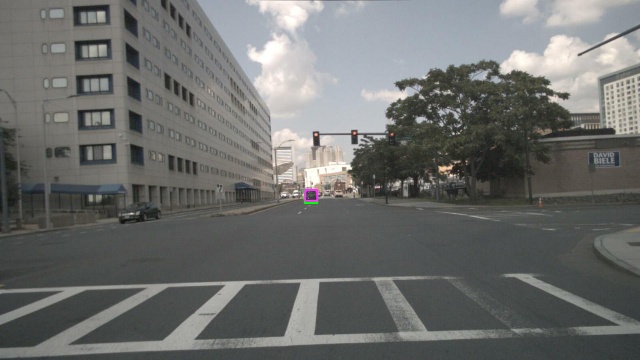}
    \end{minipage}
    & 
    \begin{minipage}{0.45\textwidth}
\includegraphics[width=\textwidth]{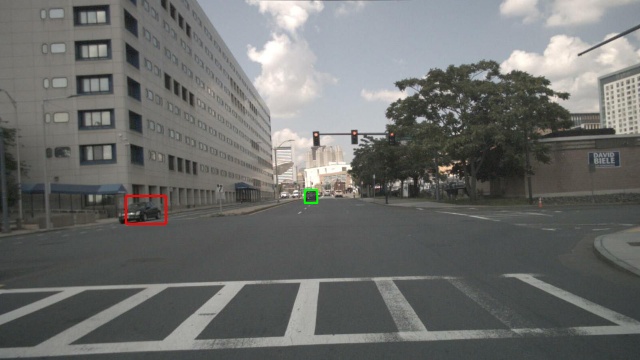}
    \end{minipage} \\
    \multicolumn{2}{c}{    \textbf{Command}: Follow \textbf{the black car ahead of me in the left lane}.
} \\
    
\end{tabular}
  \end{center}
\caption{Examples from the challenging \textbf{depth sub-test set} where the \texttt{MSRR} (left) successfully locates the referred object while \texttt{MAC} fails. For each image, the command is given together with the referred object indicated in bold in the text. In the images itself, the green bounding box indicates the ground truth, purple indicates the output of the \texttt{MSRR} and red the output bounding box of \texttt{MAC}.}
\label{fig:depth-visual-examples}
\end{figure*}

\begin{figure*}[h!]
  \begin{center}
  \begin{tabular}{c c}
    \textbf{MSRR} & \textbf{MAC} \\
    \begin{minipage}{0.45\textwidth}
      \includegraphics[width=\textwidth]{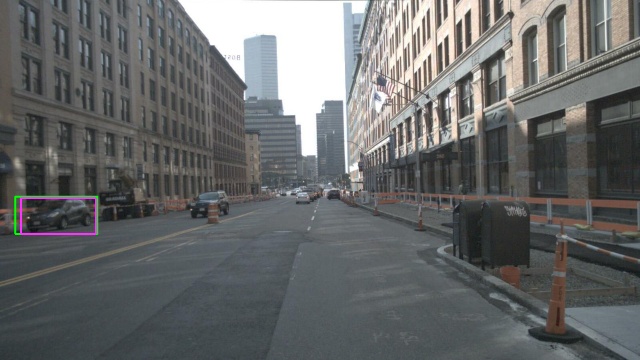}
    \end{minipage}
    & 
    \begin{minipage}{0.45\textwidth}
      \includegraphics[width=\textwidth]{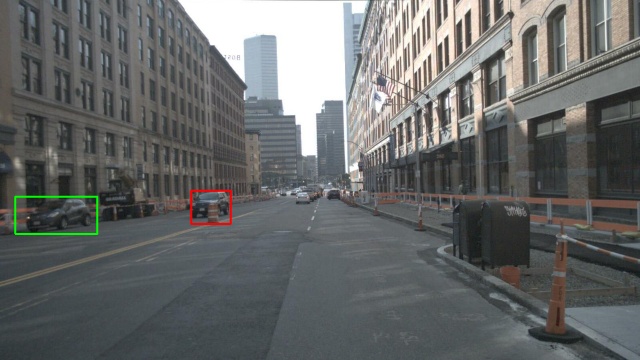}
    \end{minipage} \\
    \multicolumn{2}{C{12cm}}{\textbf{Command}: Bob is waiting in \textbf{the car on the left side of the road}. I need to pick him up. Make a you-turn.}
    \\
    \begin{minipage}{0.45\textwidth}
   \includegraphics[width=\textwidth]{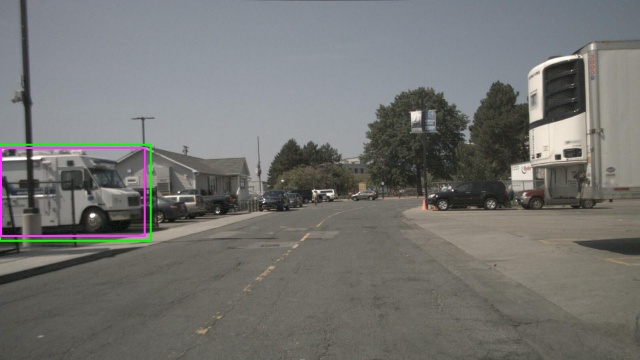}
    \end{minipage}
    & 
    \begin{minipage}{0.45\textwidth}
\includegraphics[width=\textwidth]{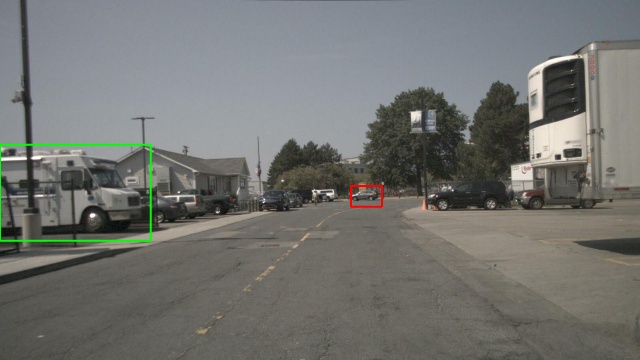}
    \end{minipage} \\
    \multicolumn{2}{C{12cm}}{\textbf{Command}: I see \textbf{Jame's truck}. It is \textbf{the one in white parked on the left}. Pull over here so I can have a chat with him.}
    \\
        \begin{minipage}{0.45\textwidth}
   \includegraphics[width=\textwidth]{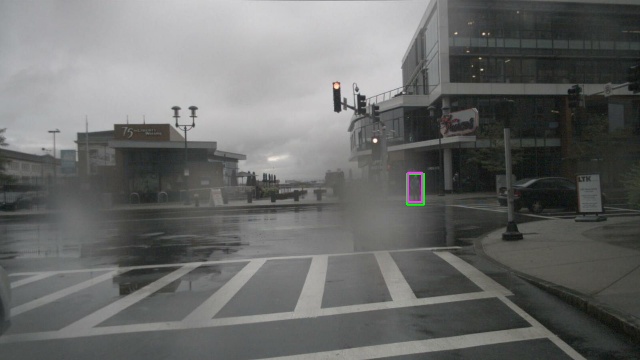}
    \end{minipage}
    & 
    \begin{minipage}{0.45\textwidth}
\includegraphics[width=\textwidth]{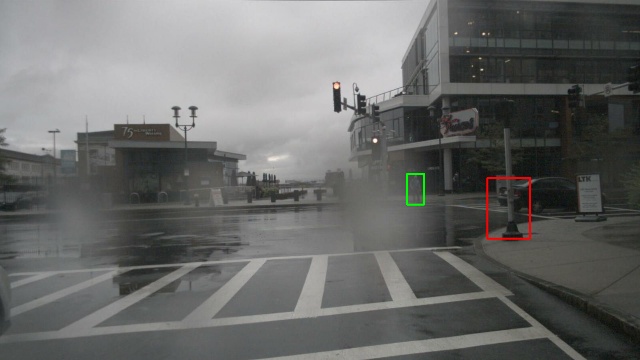}
    \end{minipage} \\
    \multicolumn{2}{C{12cm}}{ \textbf{Command}: You see \textbf{the guy at the crosspath}. I have an appointment with him. Take a right and let me out at the crosswalk.}
    
\end{tabular}
\end{center}
\caption{Examples from the challenging \textbf{long commands sub-test set} where the \texttt{MSRR} (left) can correctly find the referred object while \texttt{MAC} (right) can not. For each image, the command is given together with the referred object indicated in bold in the text. In the images itself, the green bounding box indicates the ground truth, purple indicates the output of the \texttt{MSRR} and red the output bounding box of \texttt{MAC}.}
\label{fig:long-commands-visual-examples}
\end{figure*}

\begin{figure*}[h!]
  \begin{center}
  \begin{tabular}{ c c }
    \textbf{MSRR} & \textbf{MAC} \\
    \begin{minipage}{0.45\textwidth}
      \includegraphics[width=\textwidth]{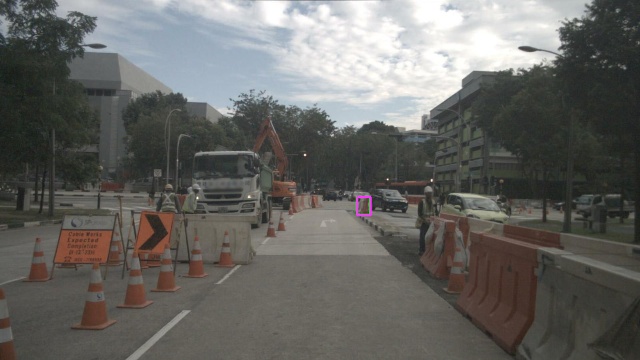}
    \end{minipage}
    & 
    \begin{minipage}{0.45\textwidth}
      \includegraphics[width=\textwidth]{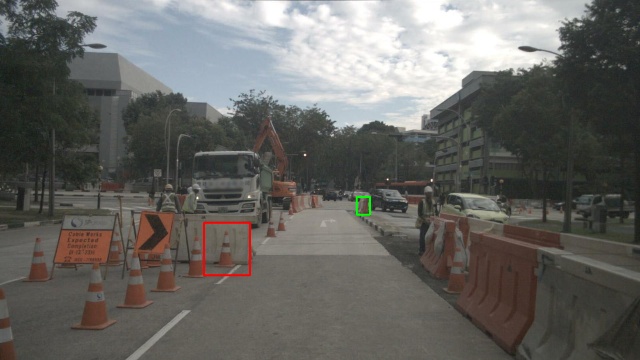}
    \end{minipage} \\
    \multicolumn{2}{c}{\textbf{Command}: Turn right around \textbf{that concrete barrier up ahead}. 
} \\
    \begin{minipage}{0.45\textwidth}
   \includegraphics[width=\textwidth]{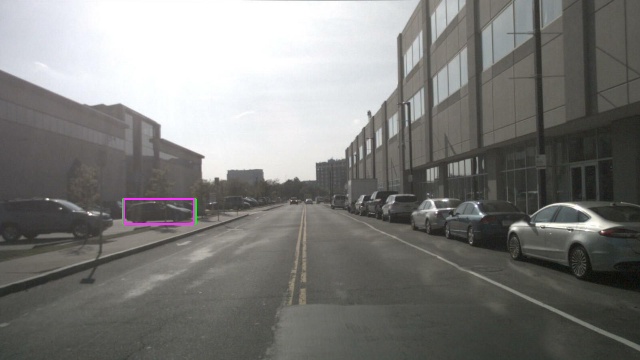}
    \end{minipage}
    & 
    \begin{minipage}{0.45\textwidth}
\includegraphics[width=\textwidth]{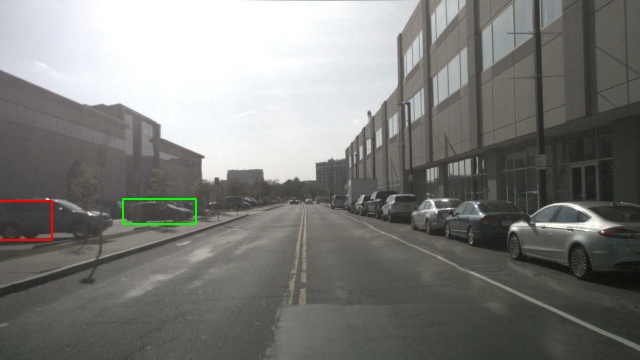}
    \end{minipage} \\
    \multicolumn{2}{c}{    \textbf{Command}: Get a parking spot near \textbf{the second car on the left side}.
} \\
        \begin{minipage}{0.45\textwidth}
   \includegraphics[width=\textwidth]{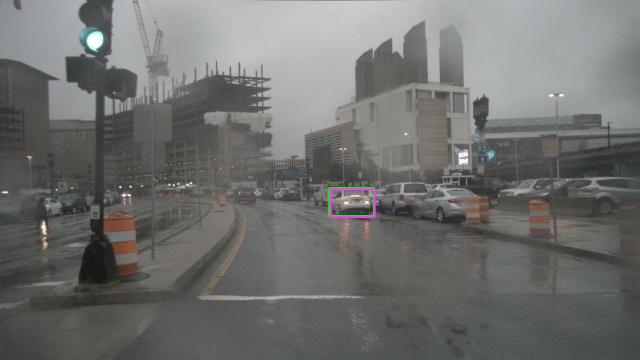}
    \end{minipage}
    & 
    \begin{minipage}{0.45\textwidth}
\includegraphics[width=\textwidth]{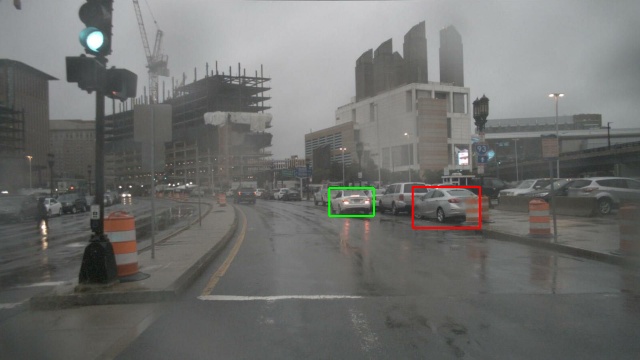}
    \end{minipage} \\
    \multicolumn{2}{c}{ \textbf{Command}: Stay in this lane to avoid \textbf{the parked car}.
}
    
\end{tabular}
\end{center}
\caption{Examples from the \textbf{ambiguity sub-test set} where the \texttt{MSRR} (left) successfully locates the referred object while \texttt{MAC} fails. For each image, the command is given together with the referred object indicated in bold in the text. In the images itself, the green bounding box indicates the ground truth, purple indicates the output of \texttt{MSRR} and red the output bounding box of \texttt{MAC}.}
\label{fig:ambiguity-visual-examples}
\end{figure*}

In this subsection we make a qualitative comparison between the current state of the art on Talk2Car, which is the \texttt{MAC} model, and our \texttt{MSRR}. We make this comparison over some images from the test set but also for images from three challenging sub-test sets provided by Talk2Car. These three sub-test sets are: depth, long commands and the ambiguity sub-test sets. 
In figure \ref{fig:success-failure-cases} we evaluate these two models on some examples from the test set. In the left column we have the output bounding box from the \texttt{MSRR} in purple while in the right column we have the output bounding box from \texttt{MAC} in red. The ground-truth bounding box of the referred object is visible in green in all images.
Under each image we also provide the referring expression, in the case of Talk2Car this is a command, where the referred object is indicated in bold.
From one of the two failure cases we see that objects that are hard to detect in the image remain a problem for both models. We argue that using an extra modality as RADAR or LIDAR could help here.
In figure \ref{fig:depth-visual-examples} we show challenging cases from the depth sub-test set where our model successful grounds the referring expression while \texttt{MAC} struggles to do so.
 Figure \ref{fig:long-commands-visual-examples} displays three images from the long commands sub-test set. Here we also show some success cases produced by our model while \texttt{MAC} struggles in these cases.
 Finally, figure \ref{fig:ambiguity-visual-examples} showcases three images from the ambiguity sub-test set showing the superiority of our model over the state of the art. 

\subsection{Visual Examples of the Reasoning Process}
\begin{figure*}[h!]
  \centering
    \includegraphics[width=\textwidth]{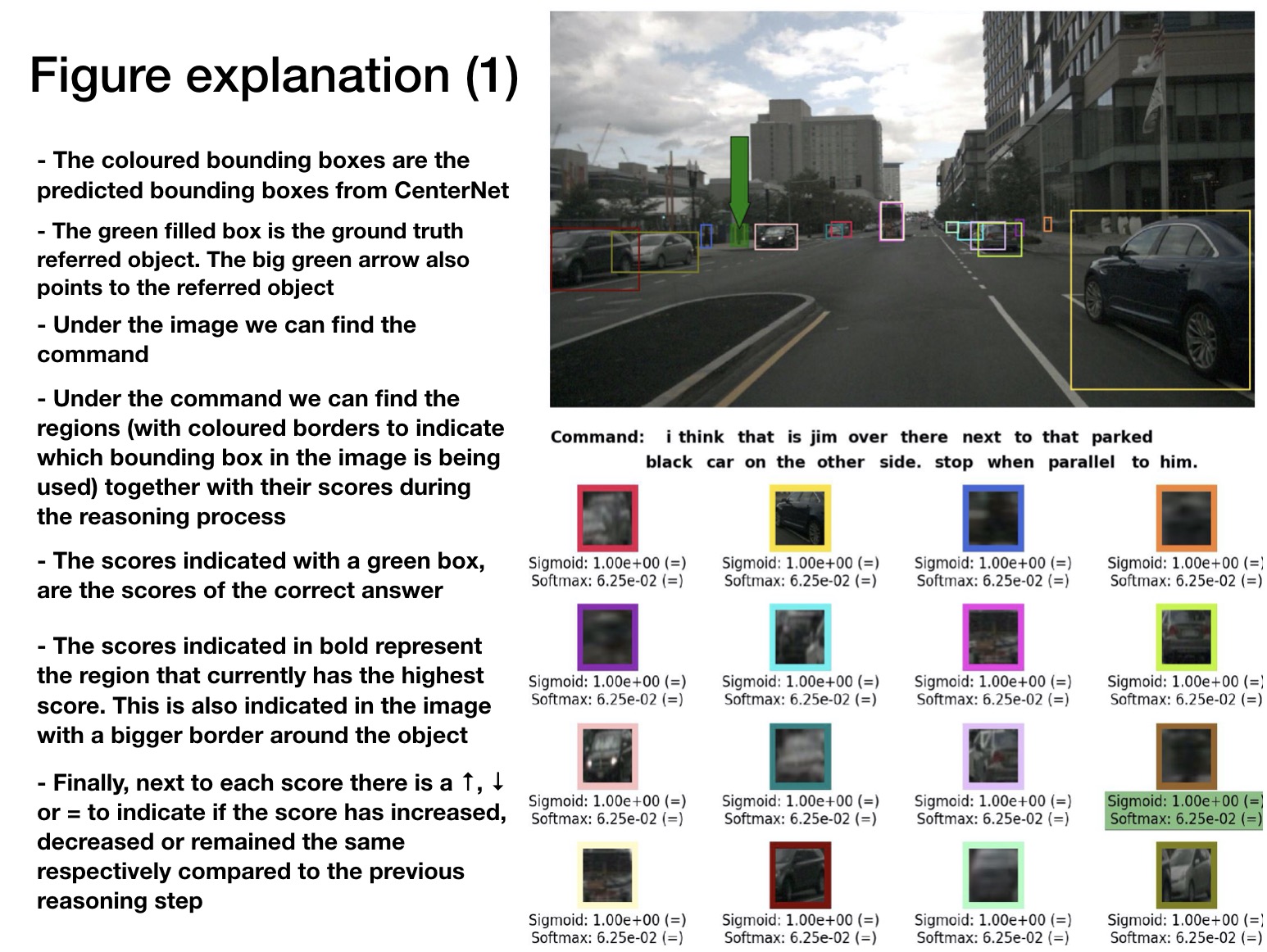}
\caption{Explaining the visualisation of the reasoning process  (Part 1).
}
    \label{fig:vis_expl1}
\end{figure*}

\begin{figure*}[h!]
  \centering
    \includegraphics[width=\textwidth]{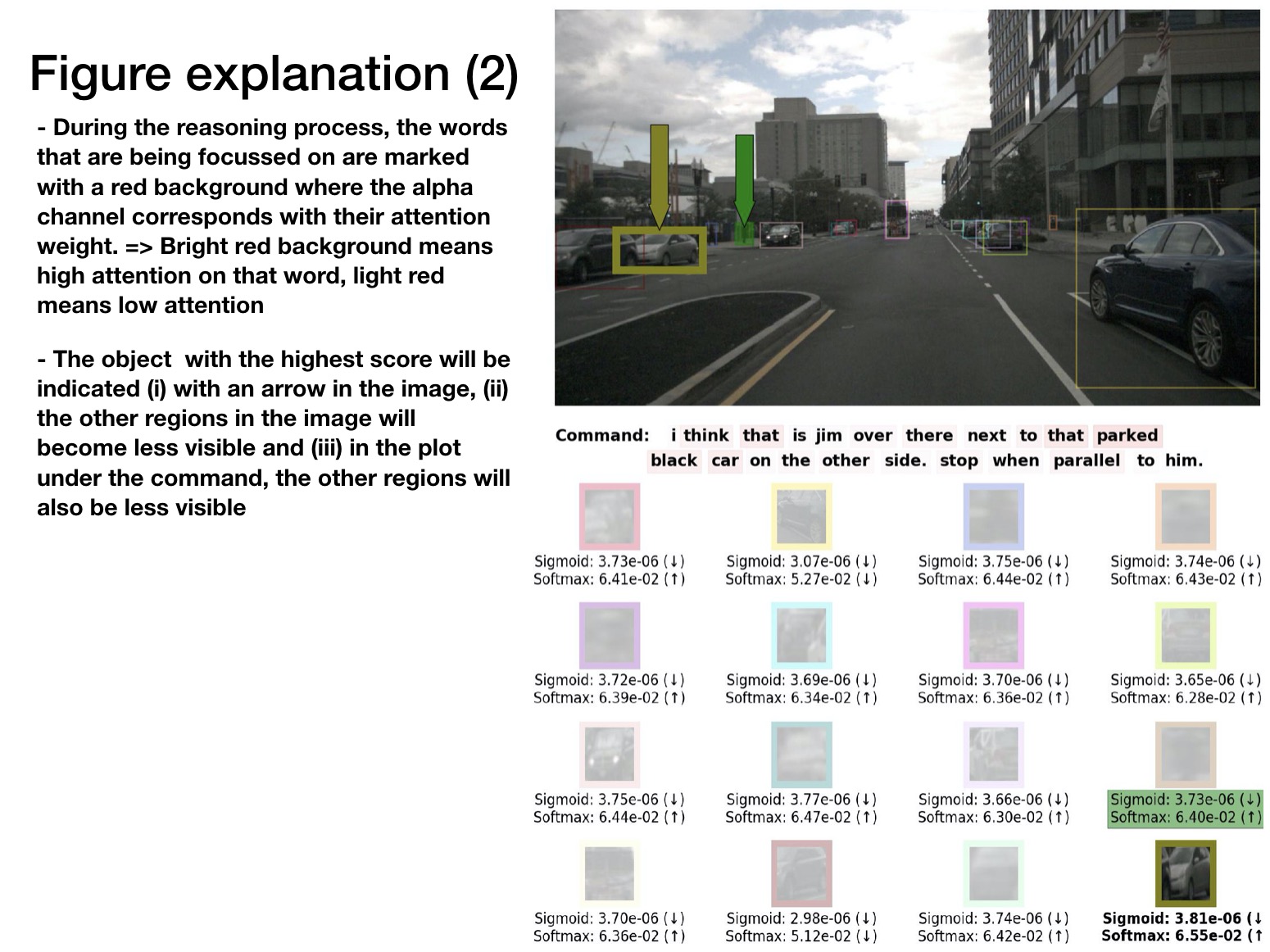}
\caption{Explaining the visualisation of the reasoning process  (Part 2).
}
    \label{fig:vis_expl2}

\end{figure*}

\begin{figure*}[h!]
  \centering
    \includegraphics[width=\textwidth]{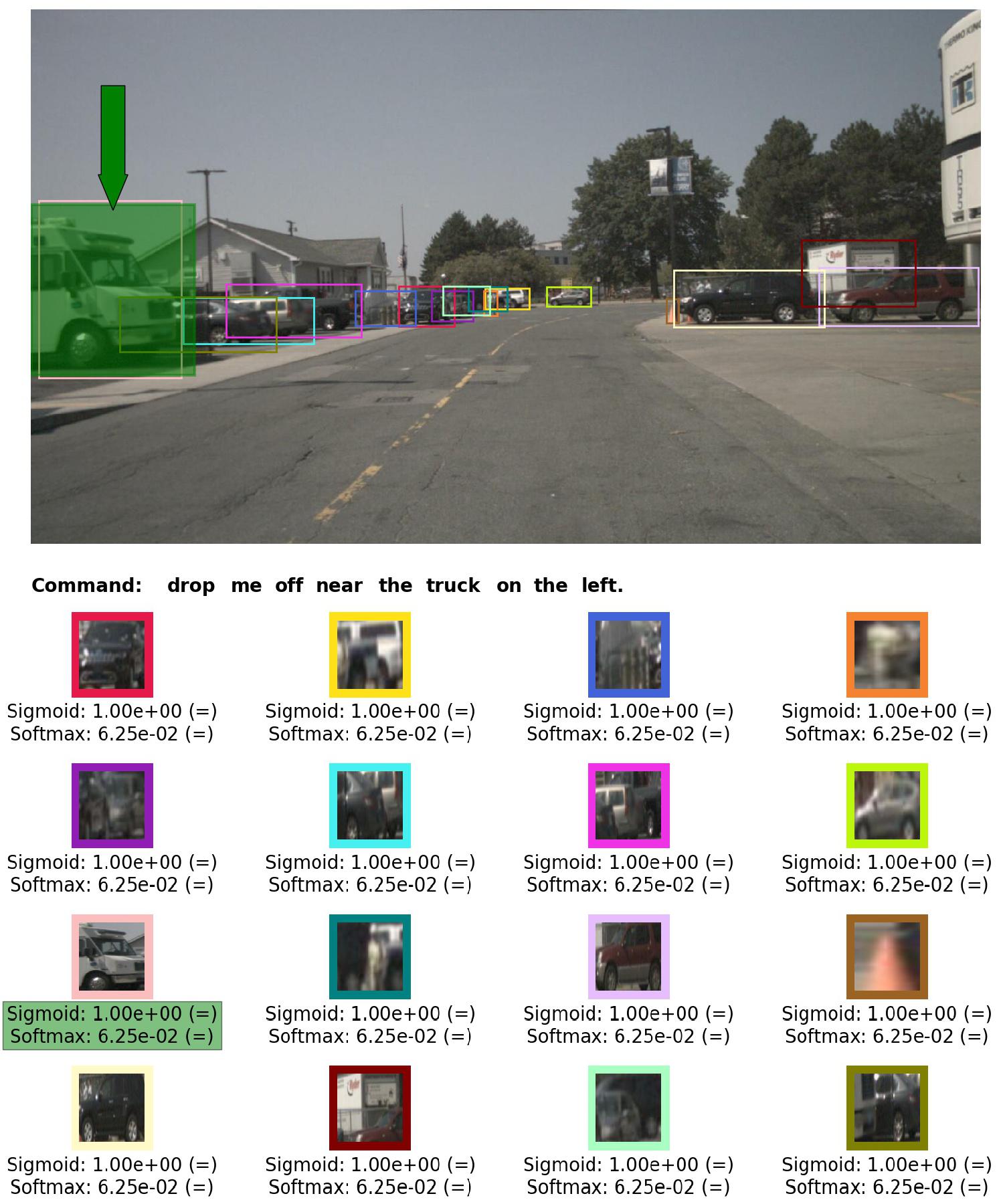}
\caption{Example 1 - The state of the model before the reasoning process starts for the given command, regions and image.}
    \label{fig:realize0}

\end{figure*}

\begin{figure*}[h!]
  \centering
    \includegraphics[width=\textwidth]{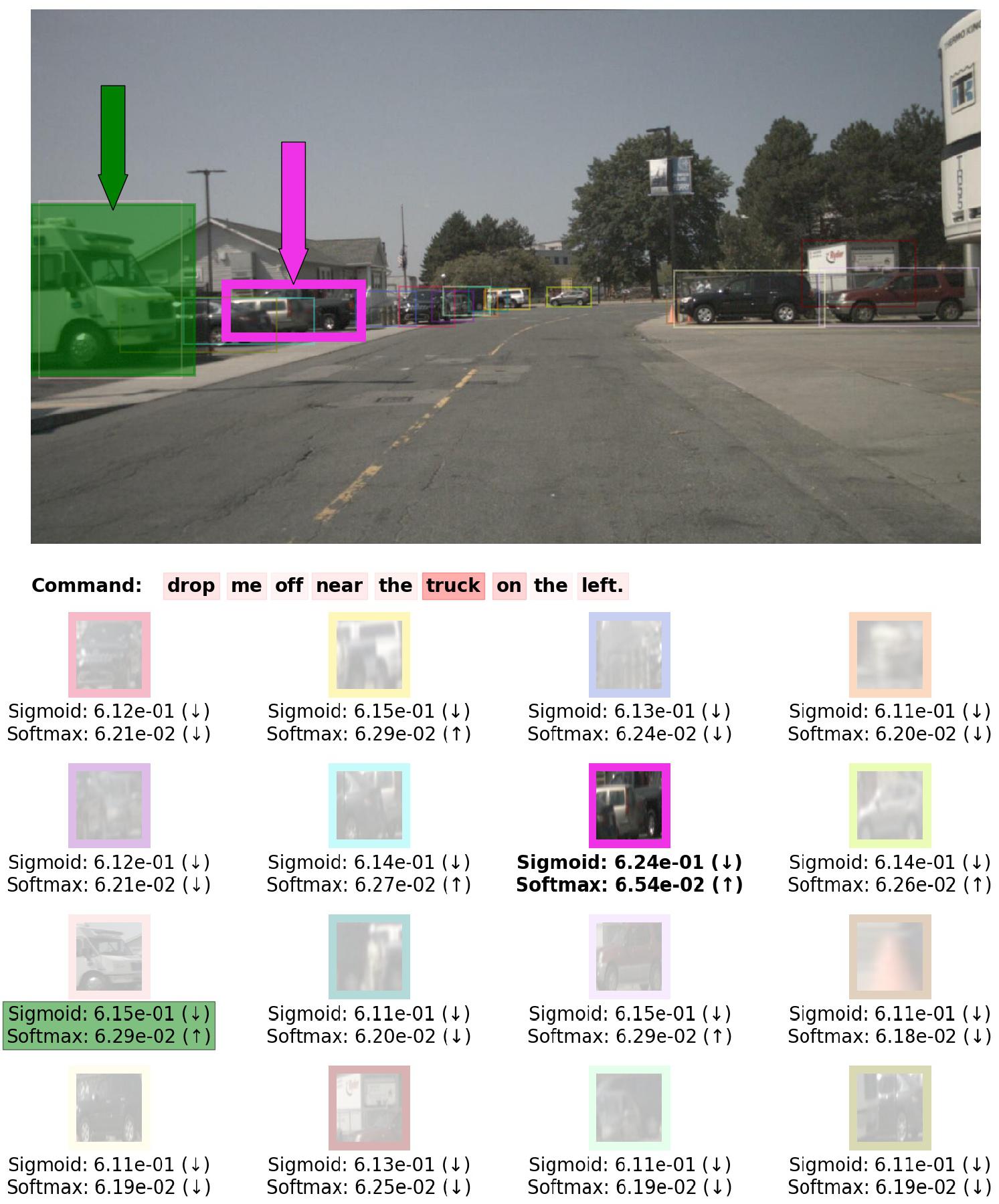}
\caption{Example 1 - Visualization of reasoning process. Step 1.}
    \label{fig:realize1}
\end{figure*}

\begin{figure*}[h!]
  \centering
    \includegraphics[width=\textwidth]{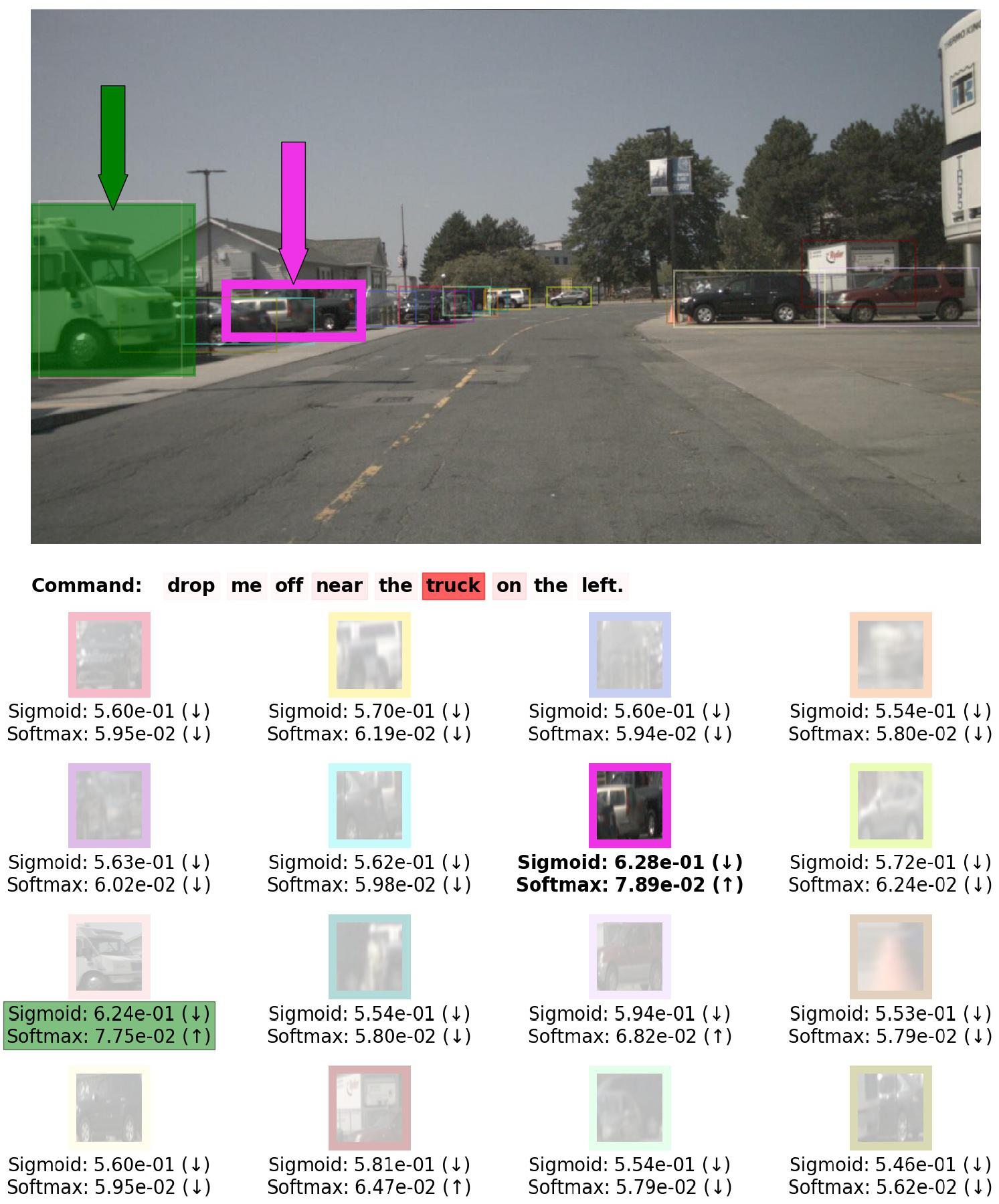}
\caption{Example 1 - Visualization of reasoning process. Step 4.}
    \label{fig:realize4}
\end{figure*}

\begin{figure*}[h!]
  \centering
    \includegraphics[width=\textwidth]{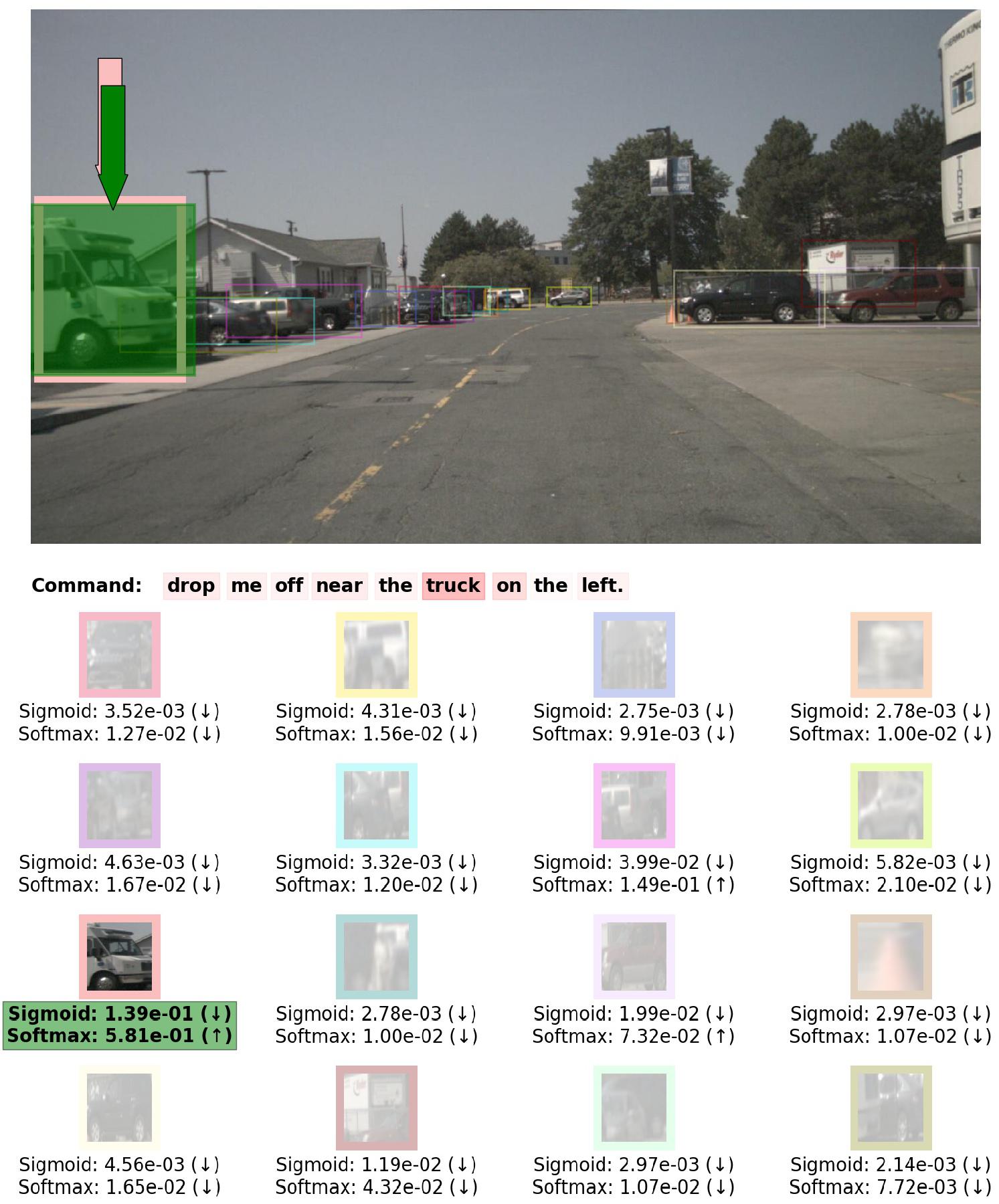}
\caption{Example 1 - Visualization of reasoning process. Final step.}
    \label{fig:realize10}
\end{figure*}

\begin{figure*}[h!]
  \centering
    \includegraphics[width=\textwidth]{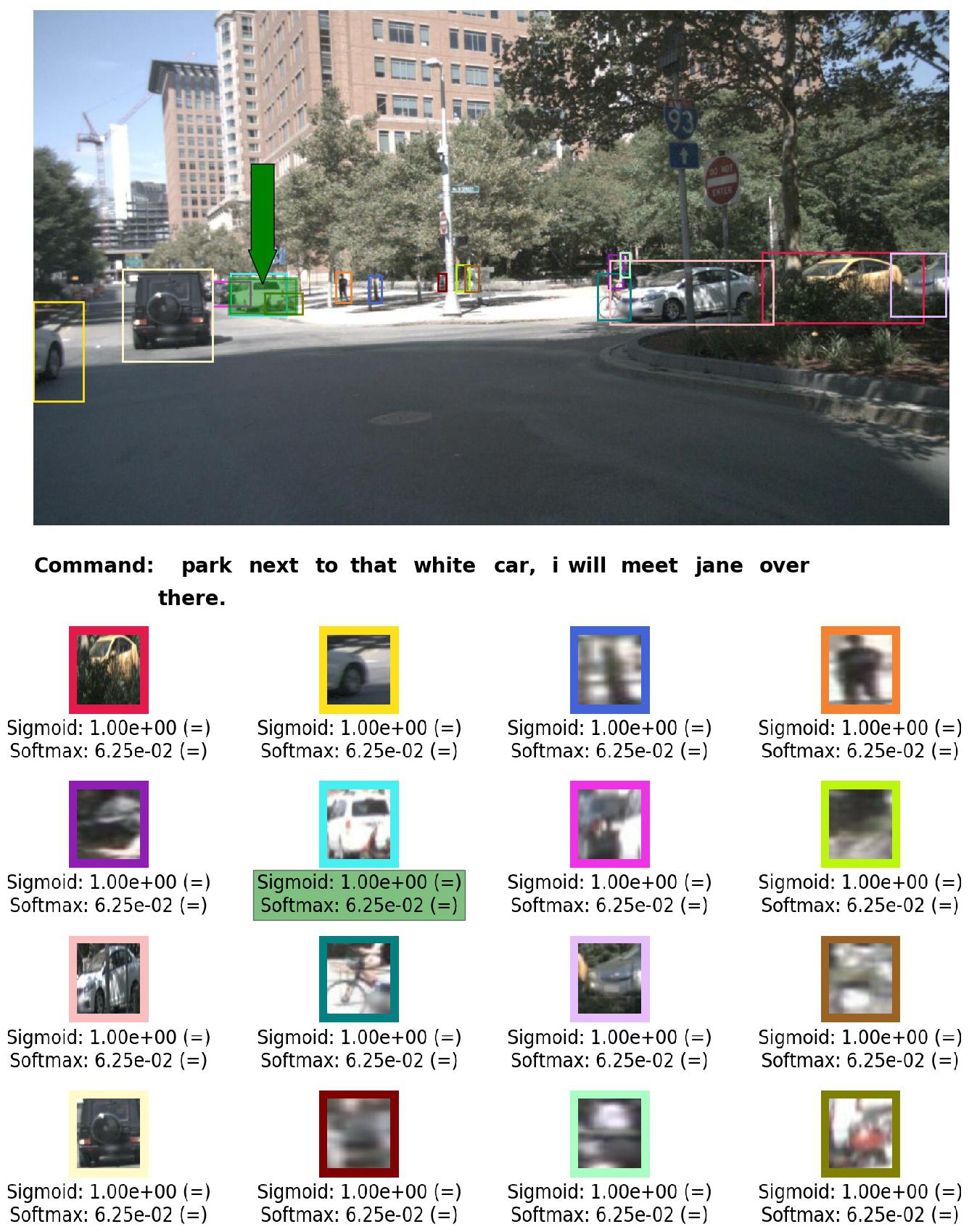}
\caption{Example 2 -  The state of the model before the reasoning process starts for the given command, regions and image.
}
    \label{fig:no-info0}
\end{figure*}

\begin{figure*}[h!]
  \centering
    \includegraphics[width=\textwidth]{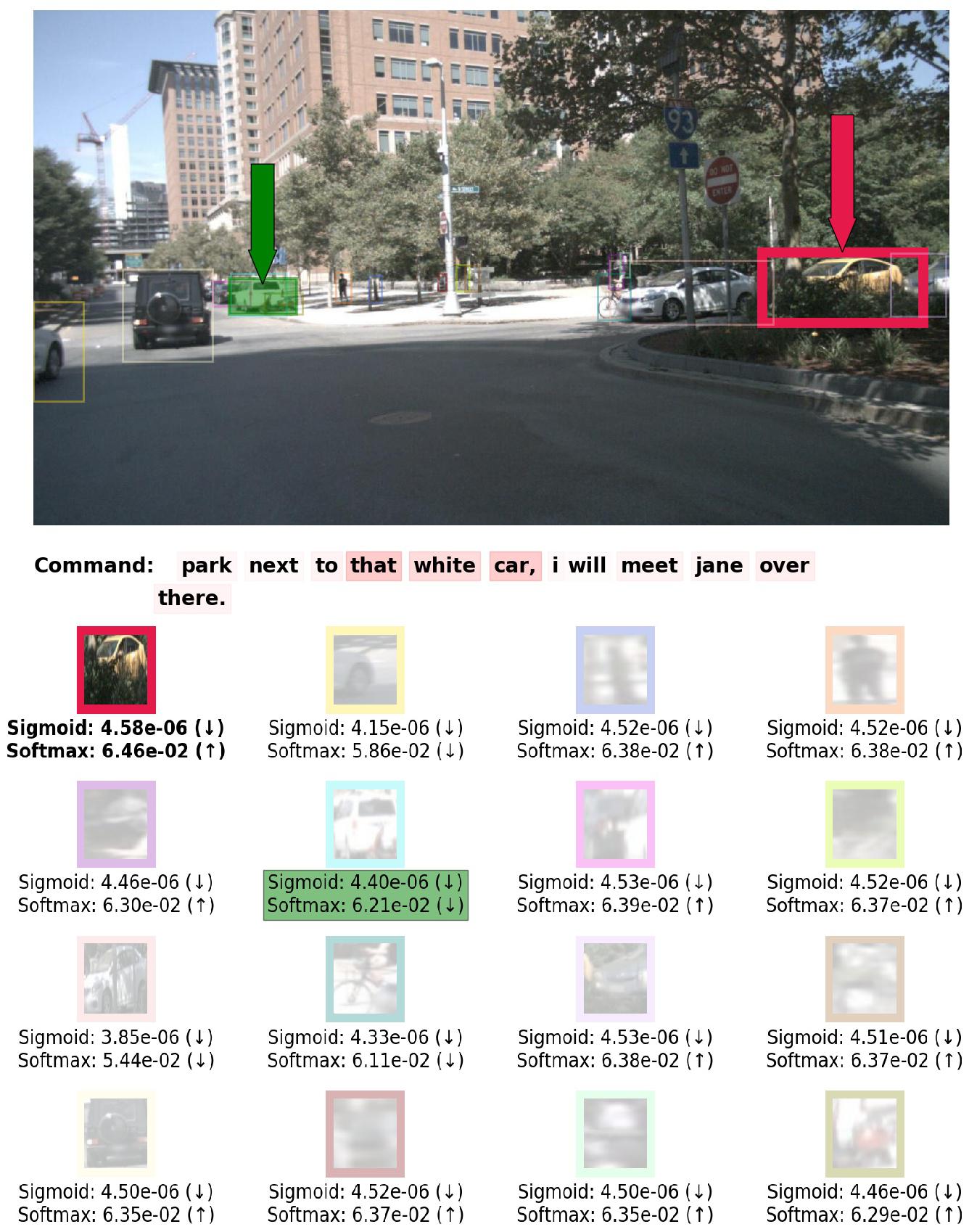}
\caption{Example 2 - Visualization of reasoning process. Step 1.
}
    \label{fig:no-info1}
\end{figure*}

\begin{figure*}[h!]
  \centering
    \includegraphics[width=\textwidth]{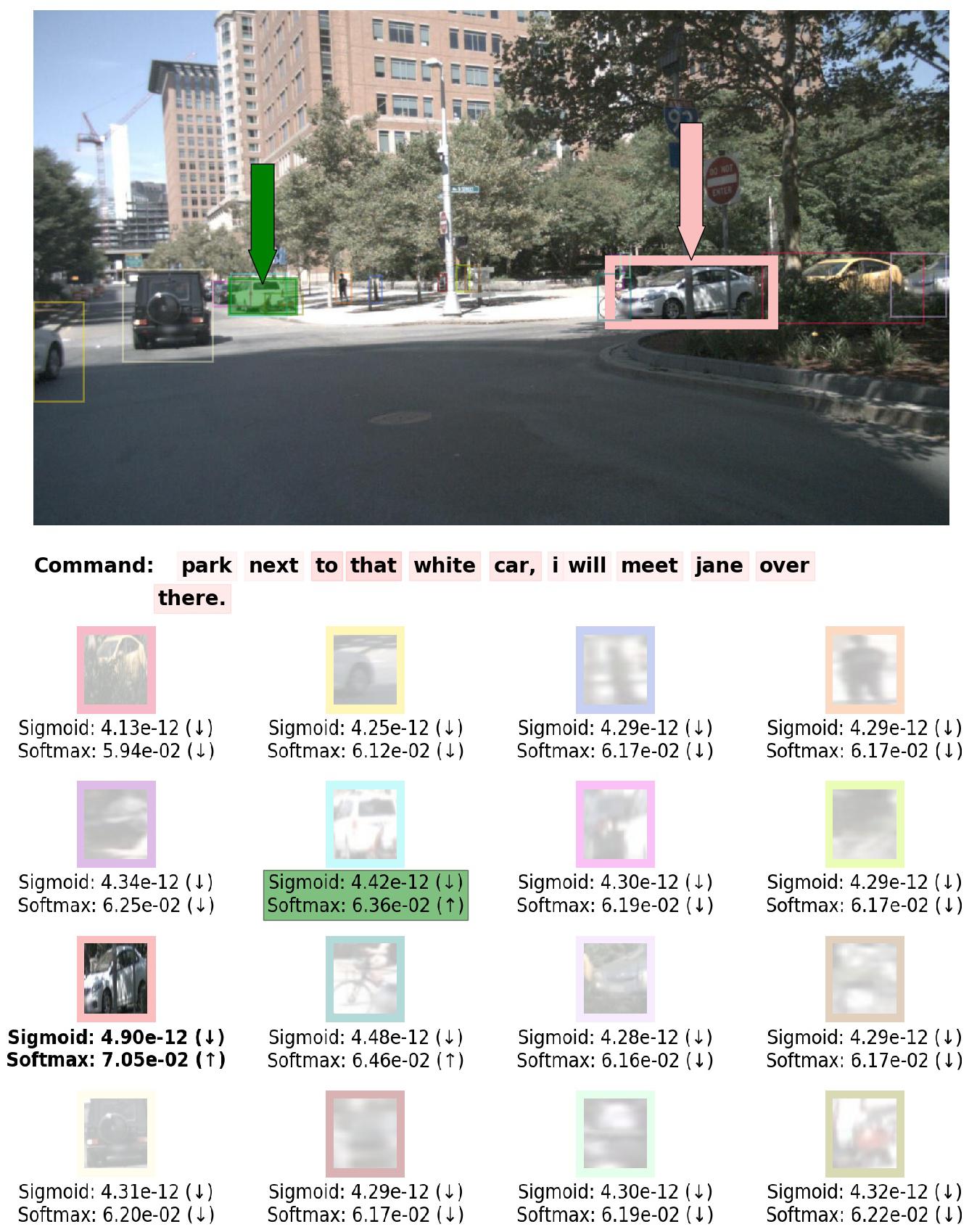}
\caption{Example 2 - Visualization of reasoning process. Step 6.
}
    \label{fig:no-info6}
\end{figure*}

\begin{figure*}[h!]
  \centering
    \includegraphics[width=\textwidth]{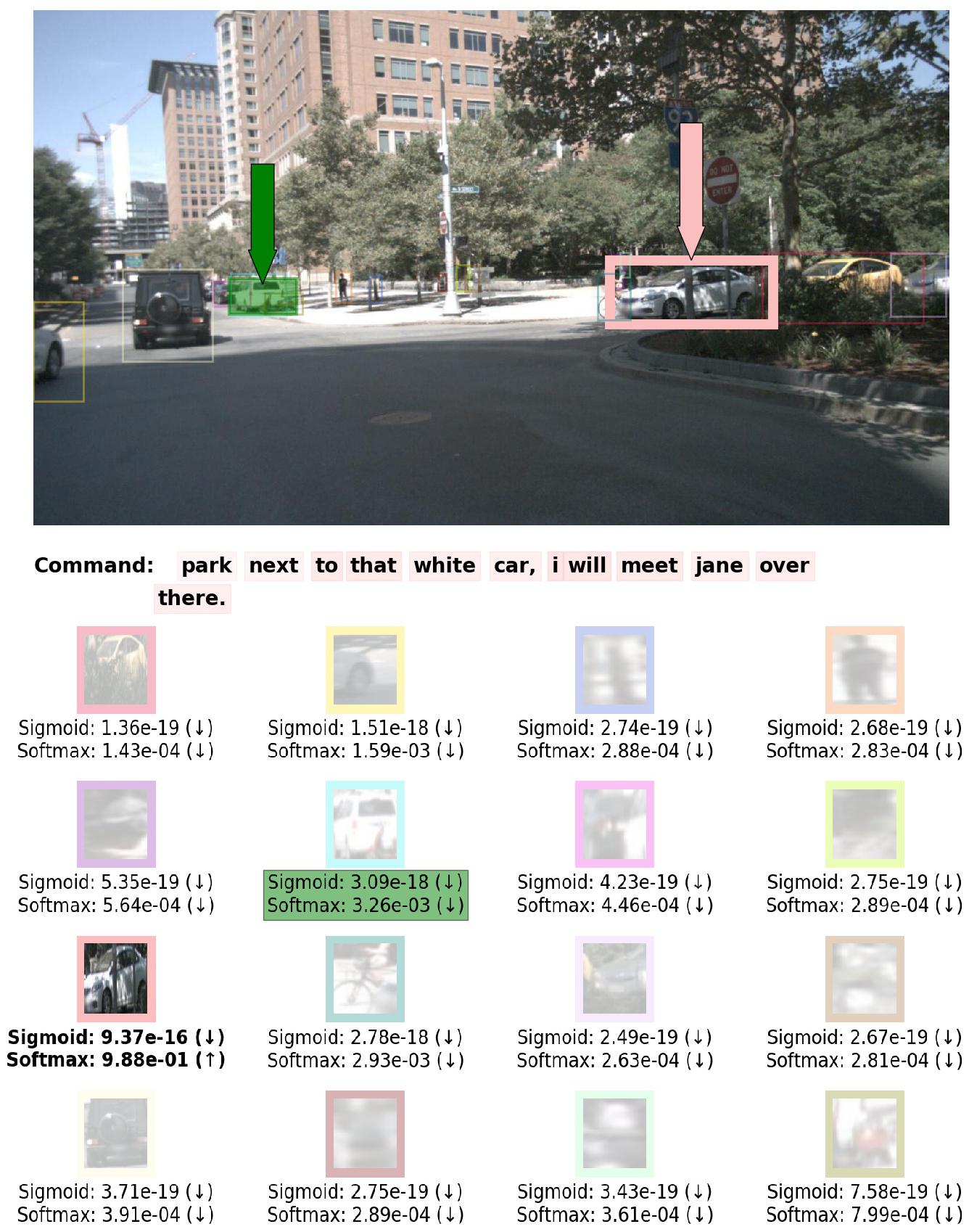}
\caption{Example 2 - Visualization of reasoning process. Final step.
}
    \label{fig:no-info10}
\end{figure*}

\begin{figure*}[h!]
  \centering
    \includegraphics[width=\textwidth]{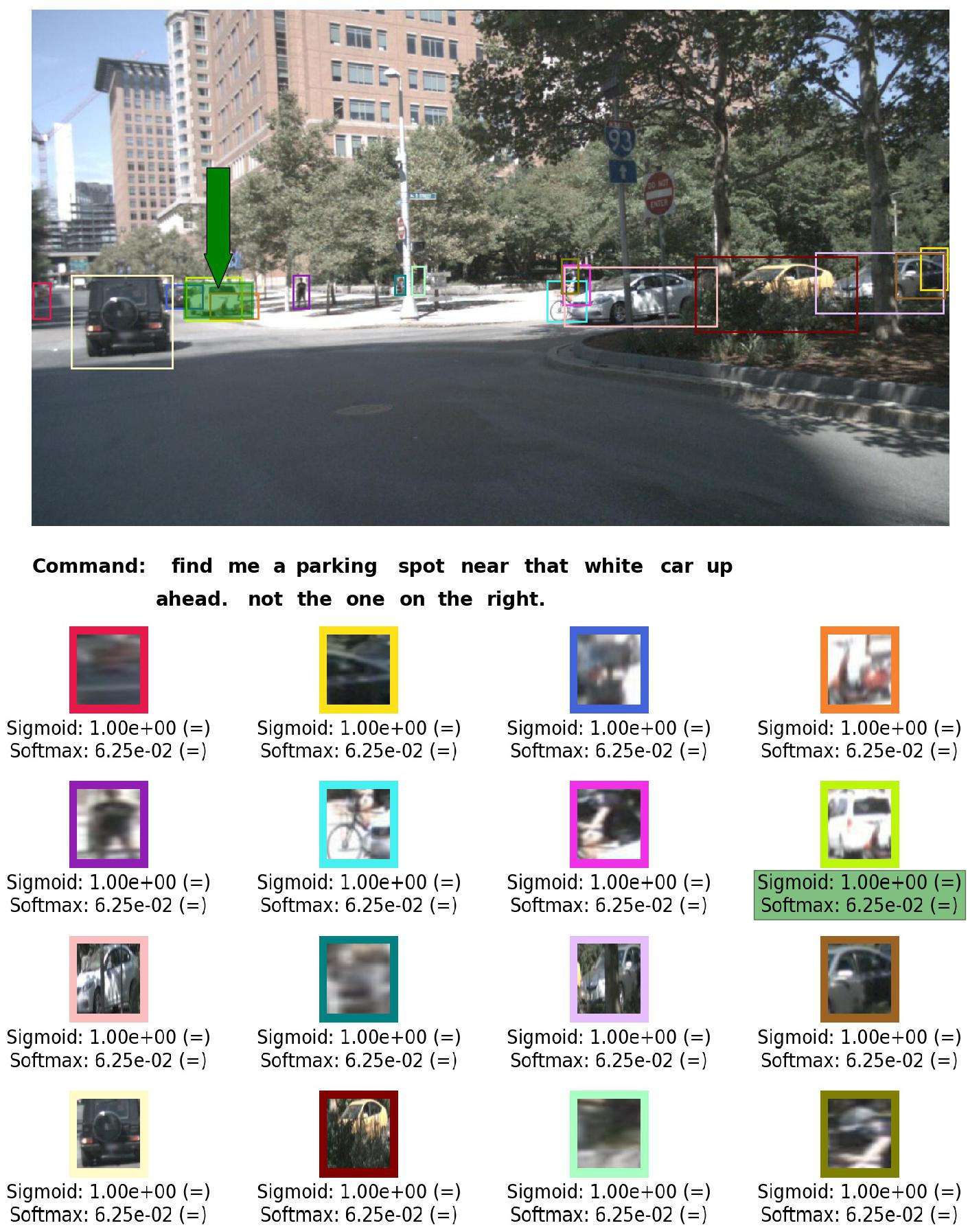}
\caption{Example 3 -  The state of the model before the reasoning process starts for the given command, regions and image.
}
    \label{fig:enough-info0}
\end{figure*}

\begin{figure*}[h!]
  \centering
    \includegraphics[width=\textwidth]{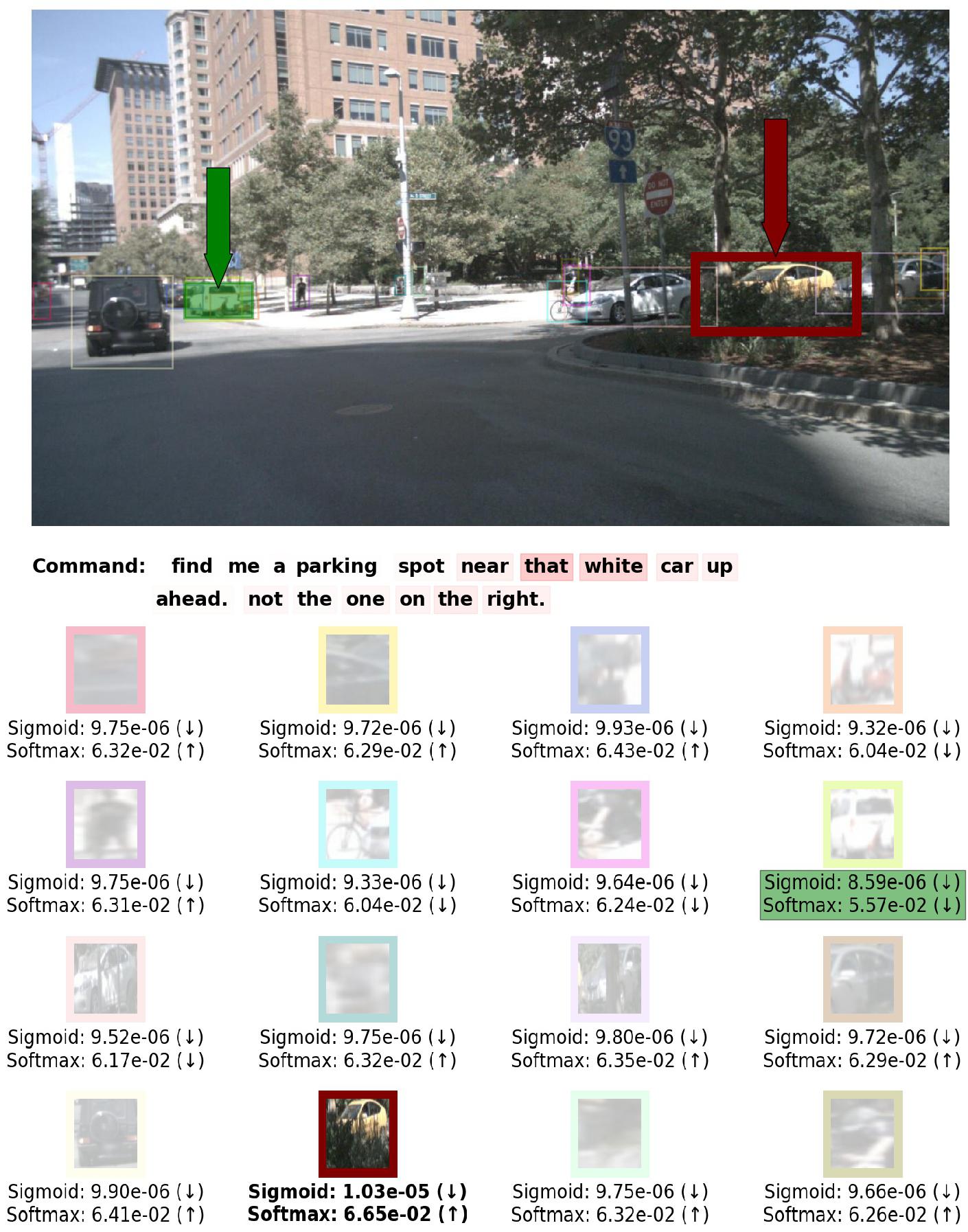}
\caption{Example 3 - Visualization of reasoning process. Step 1.
}
    \label{fig:enough-info1}
\end{figure*}

\begin{figure*}[h!]
  \centering
    \includegraphics[width=\textwidth]{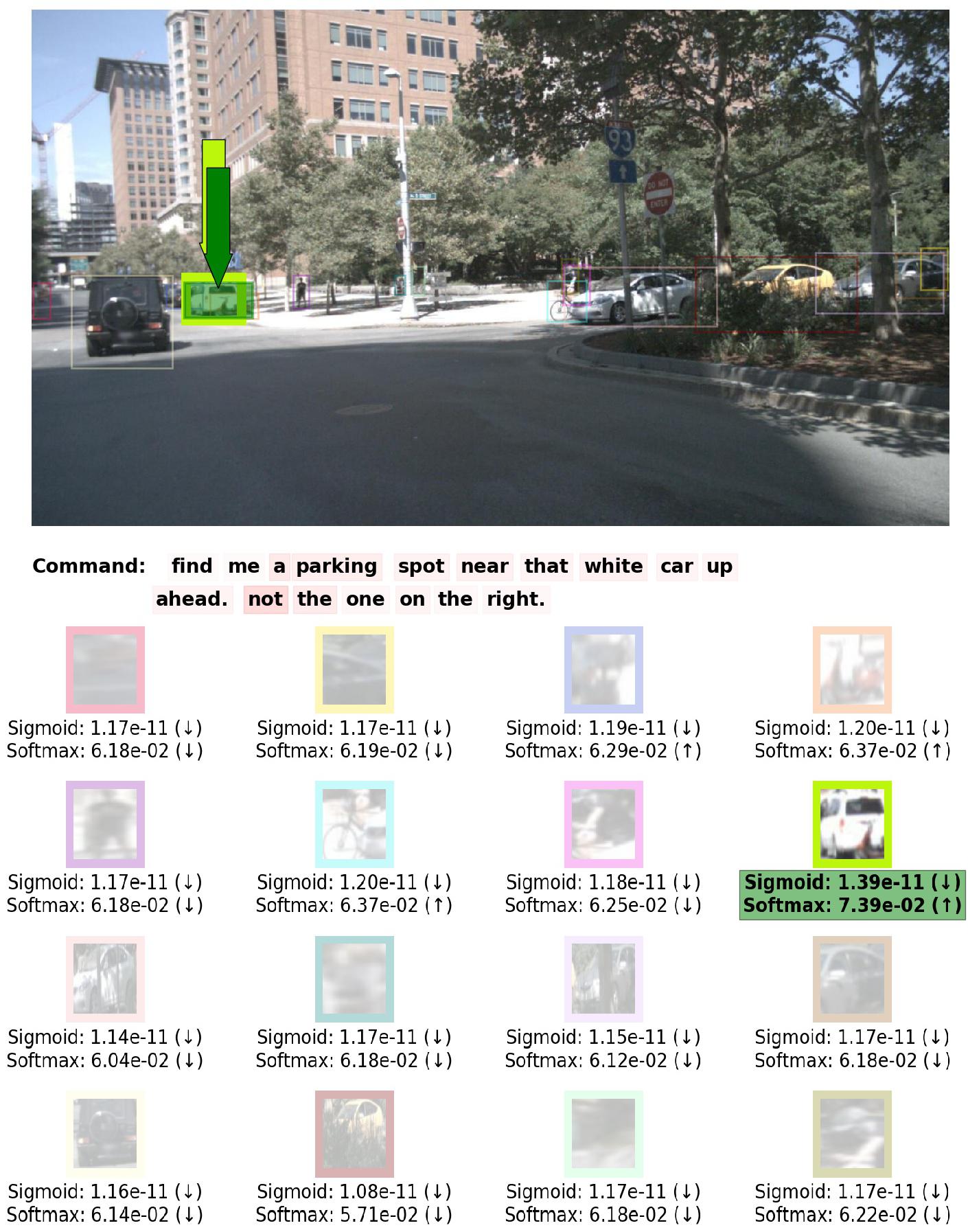}
\caption{Example 3 - Visualization of reasoning process. Step 6.
}
    \label{fig:enough-info6}
\end{figure*}

\begin{figure*}[h!]
  \centering
    \includegraphics[width=\textwidth]{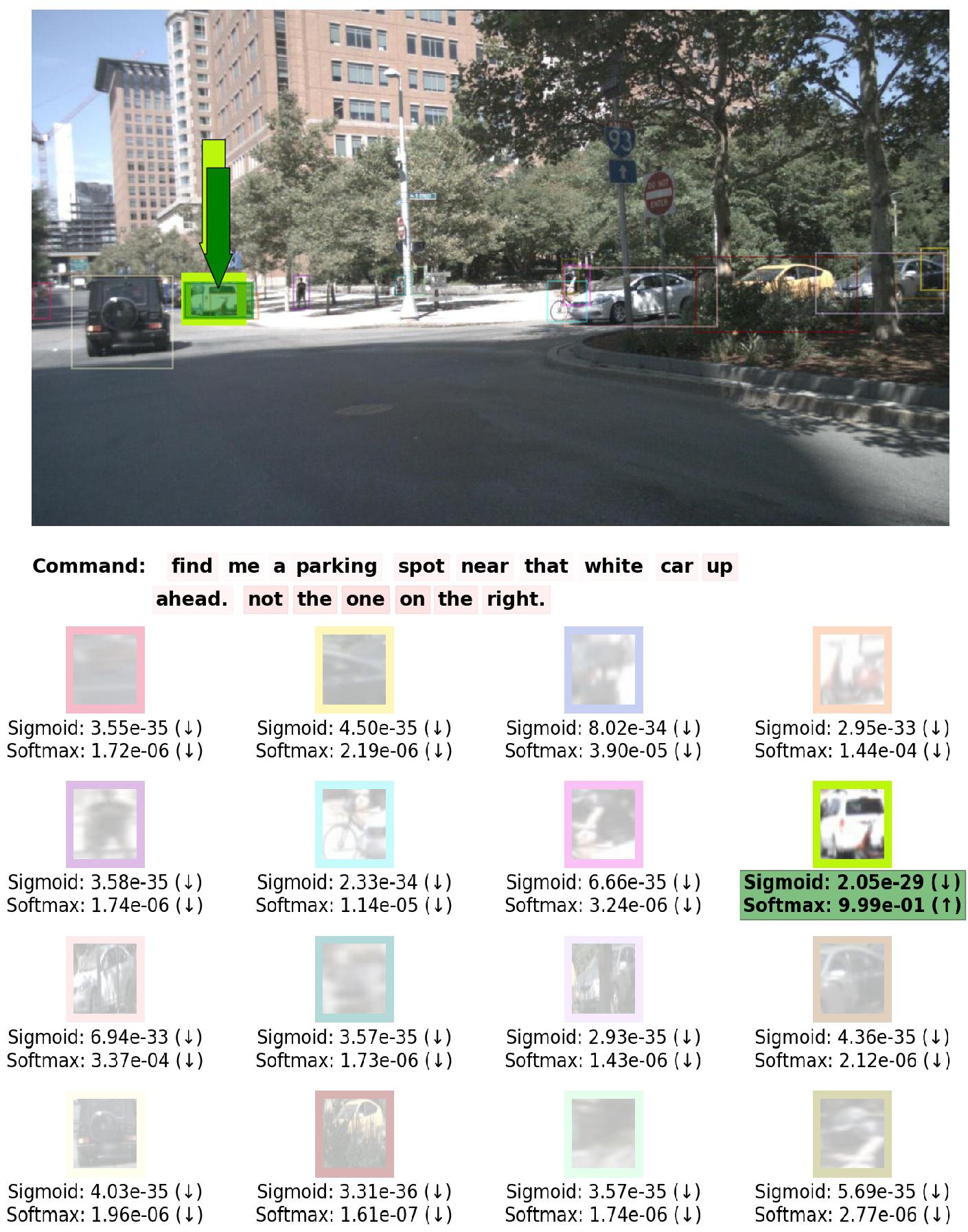}
\caption{Example 3 - Visualization of reasoning process. Final step.
}
    \label{fig:enough-info10}
\end{figure*}

In this section we showcase the visualisation of the reasoning process.
Before showing the visualisation, we advise the reader to first look at the figures \ref{fig:vis_expl1} and \ref{fig:vis_expl2} as they explain our used visualisation for the reasoning process.
We visualize the reasoning process for three different commands. For brevity we only display the steps where something interesting happens.
The first command can be seen in the figures \ref{fig:realize0} to \ref{fig:realize10}.
Figure \ref{fig:realize0} shows all the values before the reasoning process starts. In figure \ref{fig:realize1} we see that the model selects the wrong object at first. But, as the reasoning process progresses, the model notices that there is actually a different object that satisfies the command better and thus switches its choice. This is done in figure \ref{fig:realize4}. The final output of the model for this command can be seen in figure \ref{fig:realize10}.
The reasoning process for the second command can be seen in the figures \ref{fig:no-info0} to \ref{fig:no-info10}. Figure \ref{fig:no-info0} shows the begin state of the network.
 In figure \ref{fig:no-info1} the model makes a mistake by selecting the yellow car. However, in figure \ref{fig:no-info6}, the model switches to a white car. Sadly, this is not the correct white car and the model stays with this decision until its last step as can be seen in figure \ref{fig:no-info10}.
Our last example can be seen in the figures \ref{fig:enough-info0} to \ref{fig:enough-info10}. Here we have the same image as the second example, but, we have a slightly different command. The begin state can be seen in figure \ref{fig:enough-info0}. In the first steps the model makes the same mistake as in the second example by selecting the yellow car again. This can be seen in figure \ref{fig:enough-info1}. However, in figure \ref{fig:enough-info6}, the model slightly focuses on the words ``not the one on the right'' and switches to the correct white car. This is also the final prediction of the model as can be seen in figure \ref{fig:enough-info10}.
 
\end{document}